\documentclass[10pt, conference, compsocconf]{IEEEtran}
\usepackage{cite}
\usepackage[pdftex]{graphicx}
\DeclareGraphicsExtensions{.pdf,.jpeg,.png,.jpg}
\usepackage[cmex10]{amsmath}
\usepackage[tight,footnotesize]{subfigure}
\usepackage{float}
\usepackage{stfloats}
\usepackage{url}

\begin{document}

\title{Real-time deep hair matting on mobile devices}

\author{\IEEEauthorblockN{Alex Levinshtein\IEEEauthorrefmark{1},
Cheng Chang\IEEEauthorrefmark{1},
Edmund Phung\IEEEauthorrefmark{1}, 
Irina Kezele\IEEEauthorrefmark{1}, 
Wenzhangzhi Guo\IEEEauthorrefmark{1}, 
Parham Aarabi\IEEEauthorrefmark{1}\IEEEauthorrefmark{2}
}
\IEEEauthorblockA{\IEEEauthorrefmark{1}ModiFace Inc.}
\IEEEauthorblockA{\IEEEauthorrefmark{2}University of Toronto}
}

\maketitle

\begin{abstract}
Augmented reality is an emerging technology in many application domains. Among them is the beauty industry, where live virtual try-on of beauty products is of great importance. In this paper, we address the problem of live hair color augmentation. To achieve this goal, hair needs to be segmented quickly and accurately. We show how a modified MobileNet CNN architecture can be used to segment the hair in real-time. Instead of training this network using large amounts of accurate segmentation data, which is difficult to obtain, we use crowd sourced hair segmentation data. While such data is much simpler to obtain, the segmentations there are noisy and coarse. Despite this, we show how our system can produce accurate and fine-detailed hair mattes, while running at over 30 fps on an iPad Pro tablet.
\end{abstract}


\section{Introduction}
\begin{figure}[!t]
	\centerline{
		\subfigure[]{\includegraphics[width=0.15\textwidth]{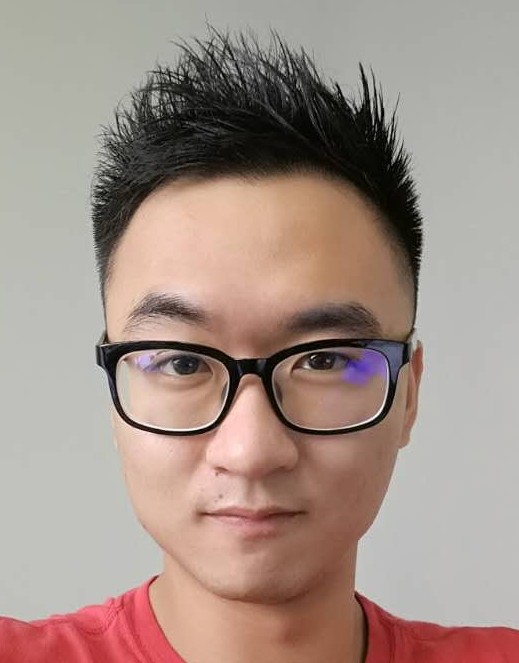}}
		\subfigure[]{\includegraphics[width=0.15\textwidth]{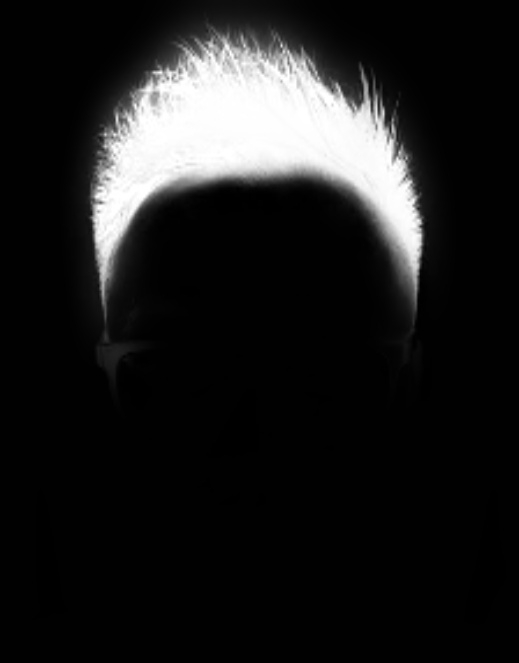}}
		\subfigure[]{\includegraphics[width=0.15\textwidth]{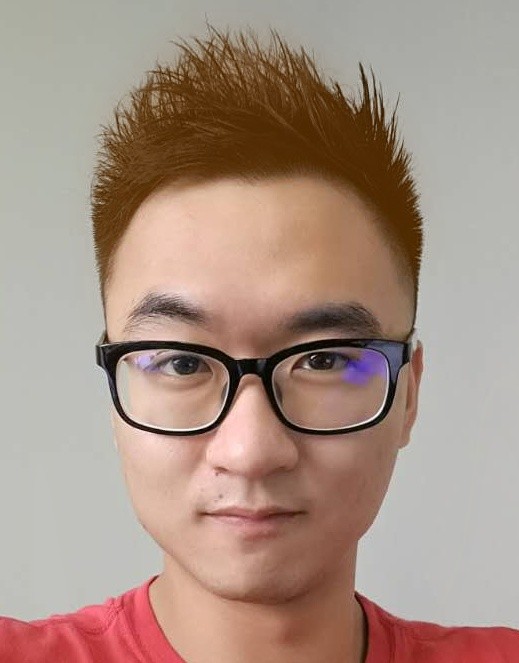}}}
	\caption{Automatic hair matting and coloring. (a) Input image. (b) Output hair matte produced by our method. (c) Recolored hair.}
	\label{fig:intro}
\end{figure}

Real-time image segmentation is an important problem in computer vision with a multitude of applications. Among them is the segmentation of hair for live color augmentation in beauty applications (Fig.~\ref{fig:intro}). This use case, however, presents additional challenges. First, unlike many objects with simple shape, hair has a very complex structure. For realistic color augmentation, a coarse hair segmentation mask is insufficient. One needs a hair matte instead. Secondly, many beauty applications run on mobile devices or in web browsers, where powerful computing resources are not available. This makes it more challenging to achieve real-time performance. This paper addresses both these challenges and introduces a system that can accurately segment hair at over $30$ fps on a mobile device.

In line with recent success of convolutional neural networks (CNNs) for semantic segmentation, our hair segmentation methods is based on CNNs. We make two main contributions. First, most modern CNNs cannot run in real-time even on powerful GPUs and may occupy a large amount of memory. Our target is real-time performance on a mobile device. In our first contribution we show how to adapt the recently proposed MobileNets \cite{howard2017mobilenets} architecture for hair segmentation, which is both fast and compact enough to be used on a mobile device. 

In the absence of detailed hair segmentation ground truth, we train our network on noisy and coarse crowd-sourced data. A coarse segmentation result, however, is insufficient for hair color augmentation purposes. For realistic color augmentation, an accurate hair matte is needed. In our second contribution, we propose a method for obtaining accurate hair mattes in real-time without the need for accurate hair matte training data. First, we show how to modify our baseline network architecture to have the capacity for capturing fine-level details. Next, by adding a secondary loss function that promotes perceptually appealing matting results, we show that the network can be trained to yield detailed hair mattes using only coarse hair segmentation training data. We compare this approach to simple Guided Filter \cite{he2013guided} post-processing and show that it yields more accurate and sharper results. 

We evaluate our method, showing that it achieves state-of-the-art accuracy while running in real-time on a mobile device. In the remainder of the paper, we discuss related work in Sec.~\ref{sec:related}, describe our approach in Sec.~\ref{sec:approach}, evaluate our method Sec.~\ref{sec:evaluation}, and conclude in Sec.~\ref{sec:summary}.

\section{Related work}
\label{sec:related}
Similar to work on general image segmentation, hair segmentation work can be divided into two categories. The first category of approaches uses hand-crafted features for segmentation. Yacoob et al. \cite{yacoob2006detection} employ simple pixel-wise color models to classify hair. Analogous method is employed by Aarabi \cite{aarabi2015automatic}, while also making use of facial feature locations and skin information. Khan et al. \cite{khan2015multi} use more advanced features with random forests for classification. Such approaches, however, prove to be insufficiently robust for real-world applications.

For more spatially consistent segmentation results, a popular method is to formulate segmentation as random field inference. Lee's et al. \cite{lee2008markov} build a Markov Random Field over image pixels, while Huang et al. \cite{huang2008towards} build their model over superpixels instead. Wang has an alternative method \cite{wang2010compositional,wang2012good} where overlapping image patches are first segmented independently and then combined.

\begin{figure*}[!t]
	\centering
	\includegraphics[width=0.9\textwidth]{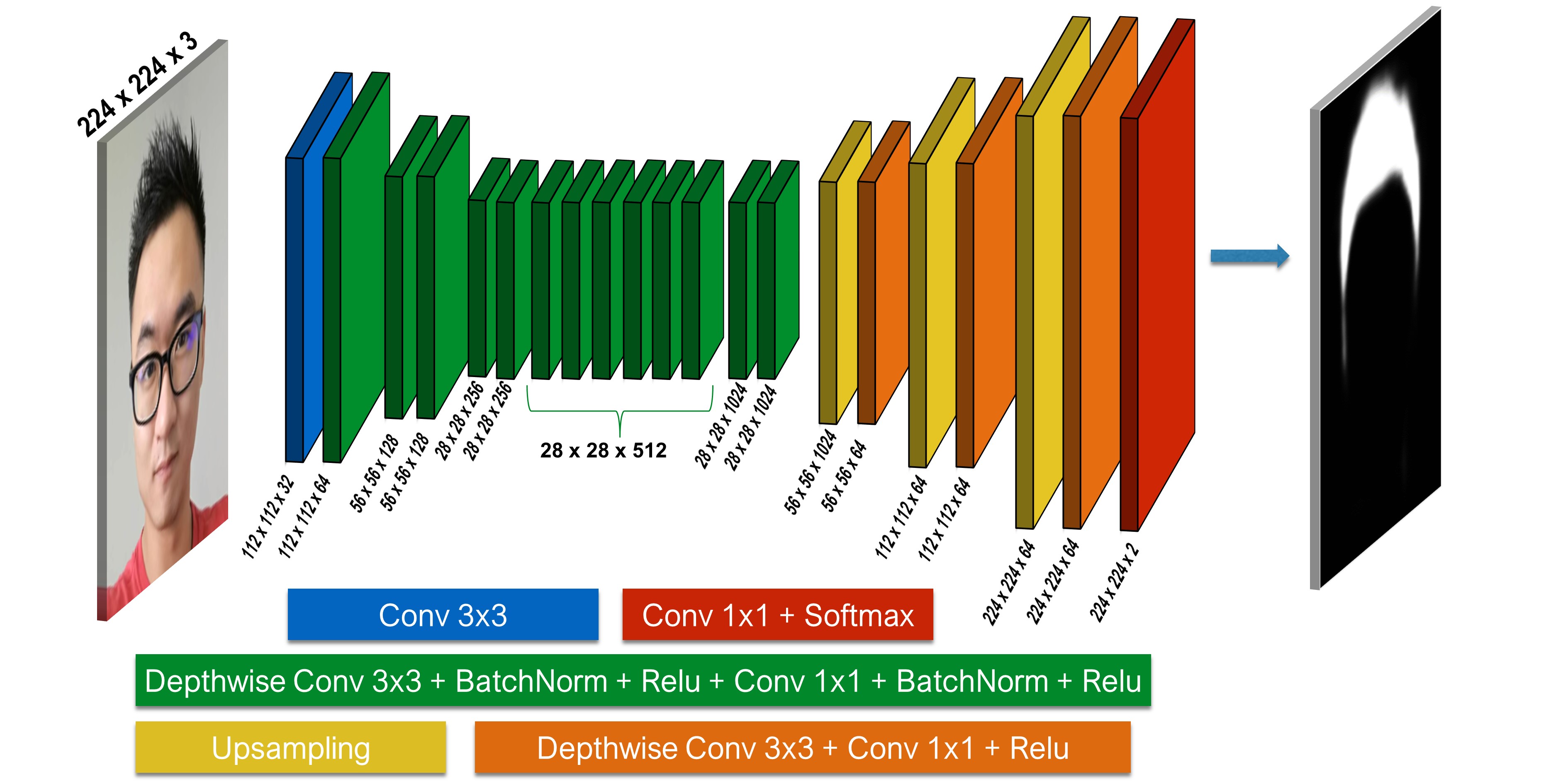}
	\caption{Fully Convolutional MobileNet Architecture for Hair Segmentation}
	\label{fig:mobilenet_segmentation}
\end{figure*}

Recently, given the success of deep neural networks (DNNs) in many areas, including semantic segmentation, DNN-based hair segmentation methods have emerged. Guo and Aarabi \cite{guo2016hair} use a heuristic method to mine high-confidence positive and negative hair patches from each image, and train a separate DNN per image, which is then used to classify the remaining pixels. Inspired by recent success of fully convolutional networks (FCN) for semantic segmentation \cite{long2015fully}, Chai et al. \cite{chai2016autohair} and Qin et al. \cite{qin2017automatic} employ FCNs for hair segmentation. Due to the coarseness of raw FCN segmentation results, similar to \cite{chen2015}, both methods post process the results using dense CRFs \cite{krahenbuhl2011efficient}. Additionally, \cite{qin2017automatic} have an extra matting step to obtain high-detail hair mattes. Finally, Xu et al. \cite{Xu_2017_CVPR} propose an end-to-end CNN architecture for generic image matting yielding state-of-the-art results. 

Our approach follows this trend, while addressing several issues. All the aforementioned methods \cite{chai2016autohair,qin2017automatic,Xu_2017_CVPR} build upon the VGG16 architecture \cite{simonyan2014very}. A single forward pass in VGG16 takes around 100ms even on a powerful GPU, and much longer on a mobile device. Adding dense CRF inference and matting futher increases the run-time. Moreover, VGG16 occupies approximately 500MB of memory, which is too much for mobile applications. Instead, we show how to adapt the recently proposed compact MobileNets architecture \cite{howard2017mobilenets} for segmentation and yield real-time matting results without expensive post processing methods. Finally, while it may be possible to obtain detailed hair matting data using semi-automatic labeling techniques \cite{Xu_2017_CVPR}, we show how to train our network without the need for such data.

\section{Approach}
\label{sec:approach}
This section describes our contributions in detail. Firstly, we describe our modifications to the original MobileNet \cite{howard2017mobilenets} architecture and the challenges of obtaining training data for hair segmentation. Secondly, we illustrate our method for real-time hair matting without the use of matting training data.

\subsection{Fully Convolutional MobileNet for Hair Segmentation}

\label{sec:approach:mobilenet}
Inspired by \cite{chai2016autohair,qin2017automatic}, we first tried to use a modified VGG16 network \cite{simonyan2014very} for hair segmentation, however a forward pass through the network took more than $2$ seconds per frame and the network occupied about $500$MB of memory. This was incompatible with our real-time mobile use case and therefore we use MobileNets \cite{howard2017mobilenets} instead, which are faster and more compact.

\begin{figure}[!t]
	\centering
	\includegraphics[width=0.45\textwidth]{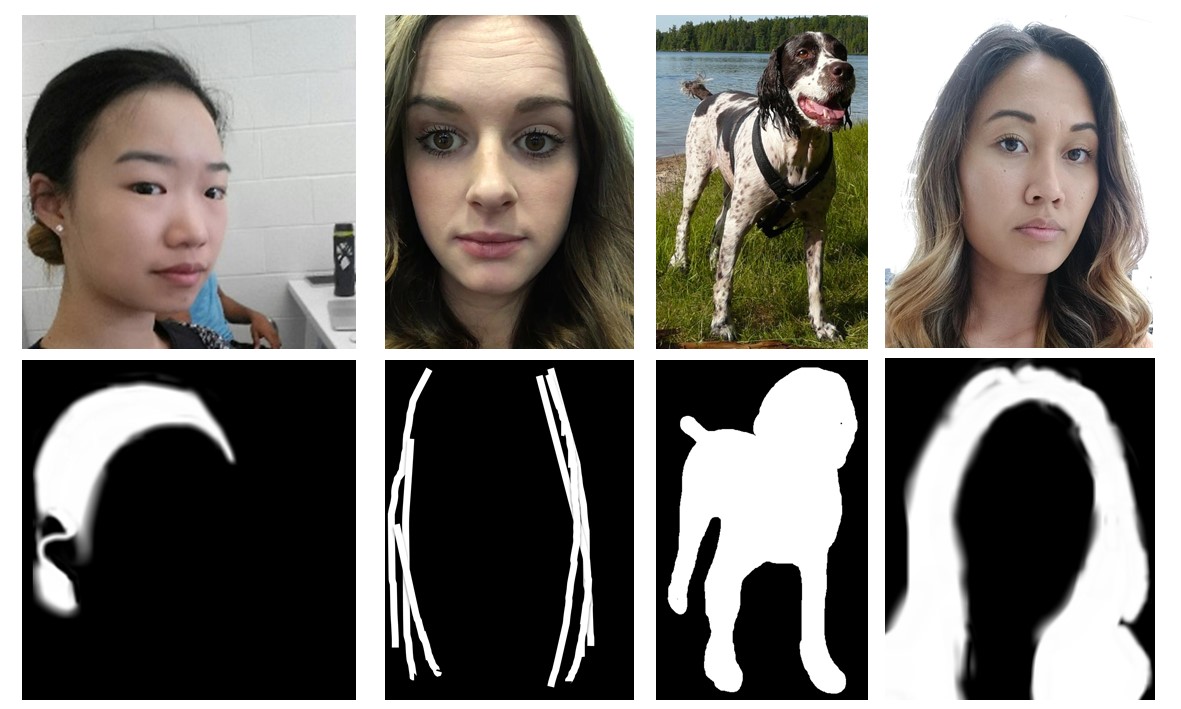}
	\caption{Crowd-sourced training data for hair segmentation. Top: images. Bottom: masks. The data is noisy and coarse, with some images having poor masks (2nd) and some non-face images (3rd).}
	\label{fig:training_data}
\end{figure}

We modified the original MobileNet architecture into a fully convolutional network for segmentation, which we name HairSegNet. First, we remove the last three layers: Avg Pool, FC, and Softmax (refer to Table 1 in \cite{howard2017mobilenets}). Next, similar to \cite{chen2015,chai2016autohair}, to preserve fine details we increase the output feature resolution by changing the step size of the last two layers with step size of $2$ to $1$. Due to our use of pre-trained weights on ImageNet, we dilate all the kernels for the layers with updated resolution by their scale factor w.r.t. their original resolution. Namely, kernels for layers that increased by a factor of $2$ are dilated by $2$ and kernels for layers that increased by a factor of $4$ are dilated by $4$. This yields a final resolution of $28\times28$.

Next, we build a decoder that takes the above CNN features as input and upsamples them to a hair mask at the original $224 \times 224$ resolution. We tried upsampling using transposed convolution layers, but saw gridding artifacts in the resulting masks. Therefore, upsampling is performed by a simplified version of an inverted MobileNet architecture. At each stage, we upsample the previous layer by a factor of $2$ by replicating each pixel in a $2\times2$ neighborhood. Then, we apply separable depthwise convolution, followed by pointwise $1\times1$ convolutions with $64$ filters, followed by ReLu. The number of filters does not have a large effect on accuracy, with $64$ filters yielding a slightly better performance based on our experiments (see Sec. \ref{sec:evaluation}). The previous block is repeated three times, yielding a $224\times224\times 64$ output. We conclude by adding a $1\times1$ convolution with softmax activation and $2$ output channels for hair / non-hair. The network is trained by minimizing the binary cross entropy loss $L_M$ between predicted and ground truth masks. The full architecture is illustrated in Fig.~\ref{fig:mobilenet_segmentation}.

The resulting architecture is considerably more compact than VGG16, occupying only $15$MB. More importantly, a forward pass takes $300$ms when implemented in Tensorflow on iPad Pro. Using the recently released optimized CoreML library \cite{coreml} from Apple, this time can be further reduced to $60$ms per frame.

Training deep neural networks requires a large amount of data. While there are large datasets for general semantic segmentation, these datasets are much less popular for hair segmentation. Moreover, unlike some objects like cars, which have a relatively simple shape, hair shape is very complex. Therefore, obtaining precise ground truth segmentation for hair is even more challenging. 

To cope with this challenge we use a pre-trained network on ImageNet and fine-tune the entire network on hair segmentation data. Nevertheless, several thousands of training images are still needed. We crowd-source such data using a hair coloring app where users have to manually mark their hair. While getting this data is cheap, the resulting hair segmentation labels are very noisy and coarse. Fig.~\ref{fig:training_data} illustrates this issue. Note that in the 2nd image hair was labeled very sparsely, while in the 3rd image a photograph of a pet was submitted. We manually clean this data by only keeping the images of human faces with sufficiently good hair masks. This is considerably faster than marking the hair from scratch or fixing incorrect segmentations.

\begin{figure*}[!t]
	\centering
	\includegraphics[width=0.9\textwidth]{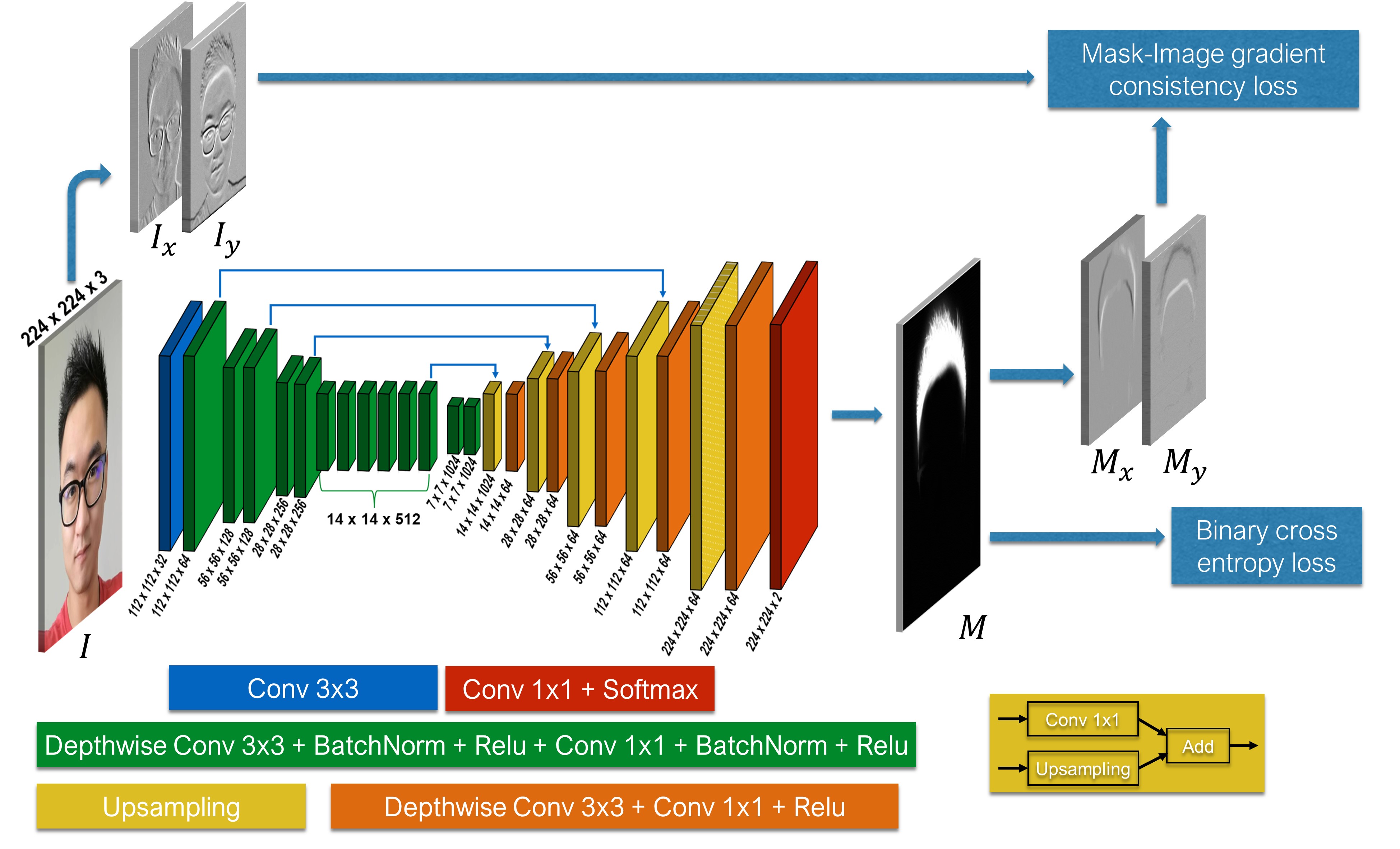}
	\caption{Fully Convolutional MobileNet Architecture for Hair Matting. Skip connections are added to increase the network capacity for capturing high resolution detail. Mask-image gradient consistency loss is added alongside with standard binary cross entropy loss to promote detailed matting results.}
	\label{fig:mobilenet_segmentation_skip}
\end{figure*}

\subsection{Hair matting}
\label{sec:approach:matting}

In our second contribution we show how to obtain accurate hair matting results. We solve the matting problem using a CNN, which we name HairMatteNet, in an end-to-end manner. Such an approach faces two challenges. First, we need an architecture with the capacity to learn high resolution matting details. The network in Sec.~\ref{sec:approach:mobilenet} may not be suitable since the results are still generated by incremental upsampling of relatively low-res layer ($28 \times 28$). Secondly, our CNN needs to learn hair matting using only coarse segmentation training data.

We address the first issue by adding skip connections between layers in the encoder and corresponding layers in the decoder, similar to many modern network architectures \cite{long2015fully}. This way, shallower layers in the encoder, which contain high-res but weak features are combined with low-res but powerful features from deeper layers. The layers are combined by first applying a $1 \times 1$ convolution to the incoming encoder layers to make the output depth compatible with the incoming decoder layers ($64$ for the three outer skip connections and $1024$ for the inner skip connection) and then merging the layers using addition. For each resolution, the deepest encoder layer at that resolution is taken for skip connection.

The second issue is addressed by adding a loss function that promotes perceptually accurate matting output. Motivated by the alpha matting evaluation work of Rhemann et al. \cite{rhemann2009perceptually}, our secondary loss measures the consistency between image and mask edges. It is minimized when the two agree. Specifically, we define our mask-image gradient consistency loss to be:

\begin{equation}
L_C = \frac{\sum{M_{mag}\big[ 1 - \left( I_x M_x + I_y M_y \right)^2 \big]}}{\sum{M_{mag}}},
\label{eqn:grad_loss}
\end{equation}

\noindent
where $(I_x, I_y)$ and $(M_x, M_y)$ are the normalized image and mask gradients respectively, and $M_{mag}$ is the mask gradient magnitude. This loss is added to the original binary cross entropy loss with a weight $w$, making the overall loss $L = L_M + w L_C$. The combination of the two losses maintains the balance between being true to training masks while generating masks that adhere to image edges. Fig.~\ref{fig:mobilenet_segmentation_skip} illustrates our new architecture and the combination of the two loss functions. 

\begin{figure}[!t]
	\centerline{
		\subfigure[]{\includegraphics[width=0.11\textwidth]{jason}}
		\subfigure[]{\includegraphics[width=0.11\textwidth]{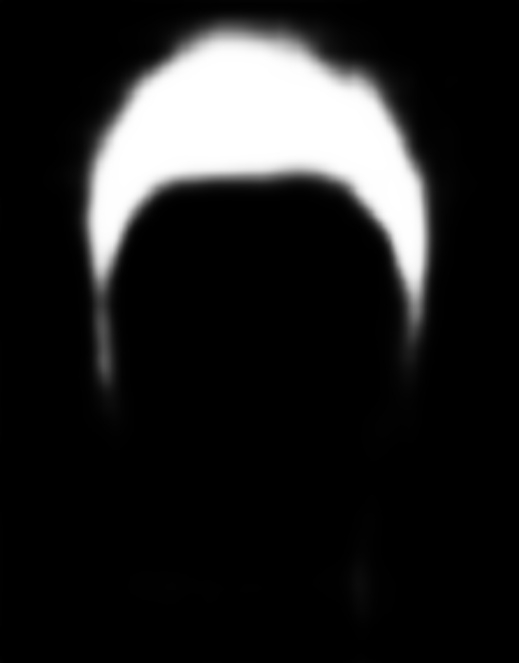}}
		\subfigure[]{\includegraphics[width=0.11\textwidth]{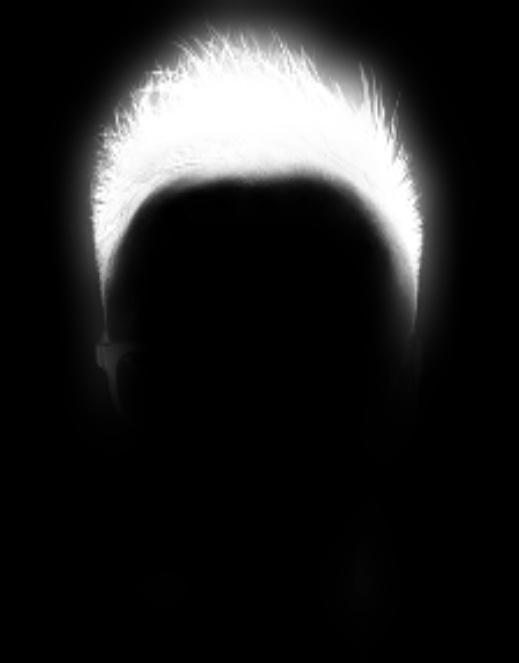}}
		\subfigure[]{\includegraphics[width=0.11\textwidth]{jason_mask_refined_CNN}}}
	\caption{Hair segmentation and matting. (a) Input image. (b) HairSegNet. (c) HairSegNet + Guided Filter. (d) HairMatteNet.}
	\label{fig:hair_matting}
\end{figure}

We compare HairMatteNet to simple post-processing of our coarse segmentation mask (HairSegNet) with a Guided Filter \cite{he2013guided}. Qin et al. \cite{qin2017automatic} used a similar approach but employed a more advanced matting method that is not fast enough for real-time applications on mobile devices. Guided Filter is an edge-preserving filter and has a linear run-time complexity w.r.t. the image size. It takes only 5ms to process a $224 \times 224$ image on iPad Pro. Fig.~\ref{fig:hair_matting} compares the masks with (c) and without (b) the filter. The former is clearly capturing more details, with individual hair strands becoming apparent. However, the filter adds detail only locally near the edges of the mask from CNN. Moreover, the edges of the refined masks have a visible halo around them, which becomes even more apparent when the hair color has lower contrast with its surroundings. This halo causes color bleeding during hair recoloring. HairMatteNet yields sharper edges (Fig.~\ref{fig:hair_matting}d) and captures longer hair strands, without the unwanted halo effect seen in Guided Filter post-processing.

As an additional bonus, HairMatteNet runs twice as fast compared to HairSegNet, taking only $30$ms per frame on a mobile device and without the need for an extra post-processing matting step. Due to our use of skip connections, that help with capturing high resolution detail, HairMatteNet maintains the original MobileNet encoder structure with the deepest layers having $7 \times 7$ resolution. These layers have many depth channels ($1024$) and become very expensive to process with increased resolution. Having a $7\times7$ resolution makes processing much faster compared to the $28\times28$ resolution in HairSegNet.

\section{Experiments}
\label{sec:evaluation}

We evaluate our method on three datasets. First is our crowd-sourced dataset, consisting of $9000$ training, $380$ validation, and $282$ testing images. All three subsets include the original images and their flipped versions. Since our target is hair matting on mobile devices, we pre-process our data by detecting the face and cropping a region around it based on the scale expected for typical selfies.

To compare our method to existing approaches, we evaluate two public datasets: LFW Parts dataset \cite{GLOC_CVPR13} and the hair dataset of Guo and Aarabi \cite{guo2016hair}. The former consists of $2927$ $250 \times 250$ images, with $1500$ training, $500$ validation, and $927$ test images. Pixels are labeled into three categories: hair, skin, and background, generated at the superpixel level. The latter consists of $115$ high-resolution images. Since it contains too few images to train on, we use our crowd-sourced training data when evaluating on this set. To make this dataset consistent with our training data, we pre-process it in a similar manner (using face detection and cropping), adding flipped images as well. Since in a few cases faces were not detected, the resulting dataset consists of $212$ images.

Training is done using a batch size of $4$ using the Adadelta \cite{zeiler2012adadelta} method in Keras \cite{chollet2015keras}, with learning rate $1.0$, $\rho=0.95$, and $\epsilon = 1e-7$. We use L2 regularization with the weight $2 \cdot 10^{-5}$ for convolution layers only. Depthwise convolution layers and the last convolution layer are not regularized. We set the loss balancing weight to $w=0.5$. In the three-class LFW data, only the hair class is contributing to the mask-image gradient consistency loss. We train our model for $50$ epochs and select the best performing epoch using validation data. Training on crowd-sourced dataset takes $5$ hours on Nvidia GeForce GTX 1080 Ti GPU and less than an hour on LFW Parts due to much smaller training set size.

\subsection{Quantitative evaluation}
For quantitative performance analysis, we measure the F1-score, Performance \cite{guo2016hair}, IoU, and Accuracy, averaged across all test images. To measure the consistency of image and hair mask edges, we also report the mask-image gradient consistency loss (Eqn.~\ref{eqn:grad_loss}). Recall that during the manual clean-up in Sec.~\ref{sec:approach:mobilenet} we only filtered images rather than correcting the masks. As a result, the quality of our hair annotation is still poor. Therefore, prior to evaluation on our crowd-sourced data, we manually corrected the test masks, spending no more than $2$ minutes per annotation. This yielded slightly better ground truth. Three variants of our method are evaluated on this relabeled data. Table~\ref{tab:crowd_eval} shows our results. All three methods perform similarly w.r.t. the ground truth comparison measures, however, HairMatteNet is the clear winner in the gradient consistency loss category, indicating that its masks adhere much better to image edges.

On the LFW Parts dataset, we report an on-par performance with the best performing method in Qin et al. \cite{qin2017automatic}, but achieve it in real-time on a mobile device. We use only the accuracy measure for evaluation since it is the only measure used in \cite{qin2017automatic}. Arguably, especially since LFW Parts was annotated at the superpixel level, the ground truth there may not good enough for high-accuracy analysis. On the dataset of Guo and Aarabi \cite{guo2016hair} we report an F1-score of $0.9376$ and Performance of $0.8253$. We re-ran HNN, the best performing method in \cite{guo2016hair}, on this post-processed dataset and obtained similar performance to that reported by the authors, with F1-score of $0.7673$ and Performance of $0.4674$.

%
\begin{figure*}[!ht]
	\centerline{
	\begin{minipage}{0.49\textwidth}
	\centerline{{\includegraphics[width=0.24\textwidth]{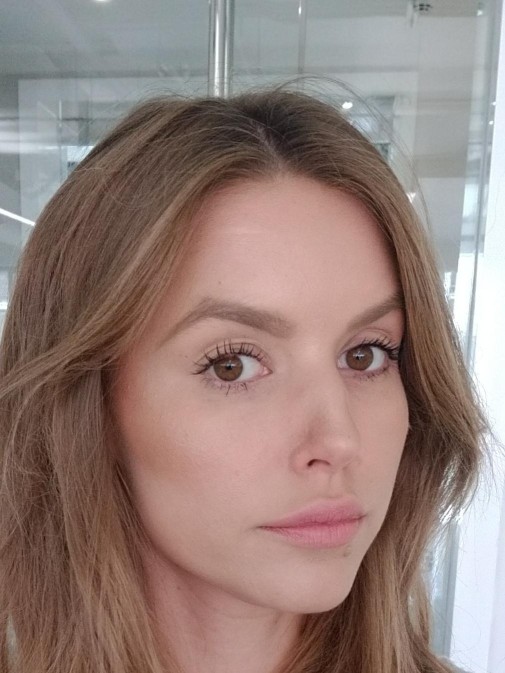}}
		{\includegraphics[width=0.24\textwidth]{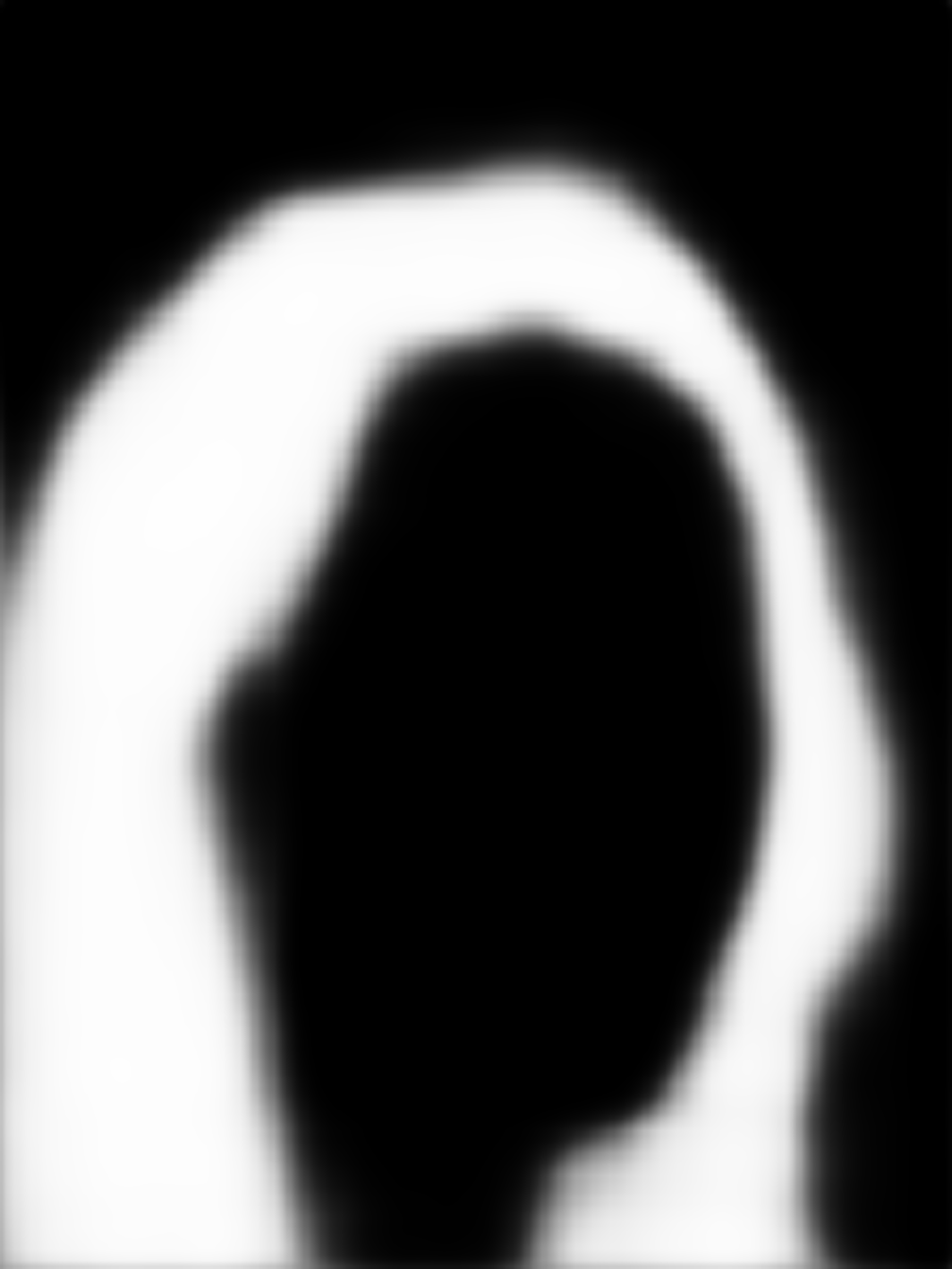}}
		{\includegraphics[width=0.24\textwidth]{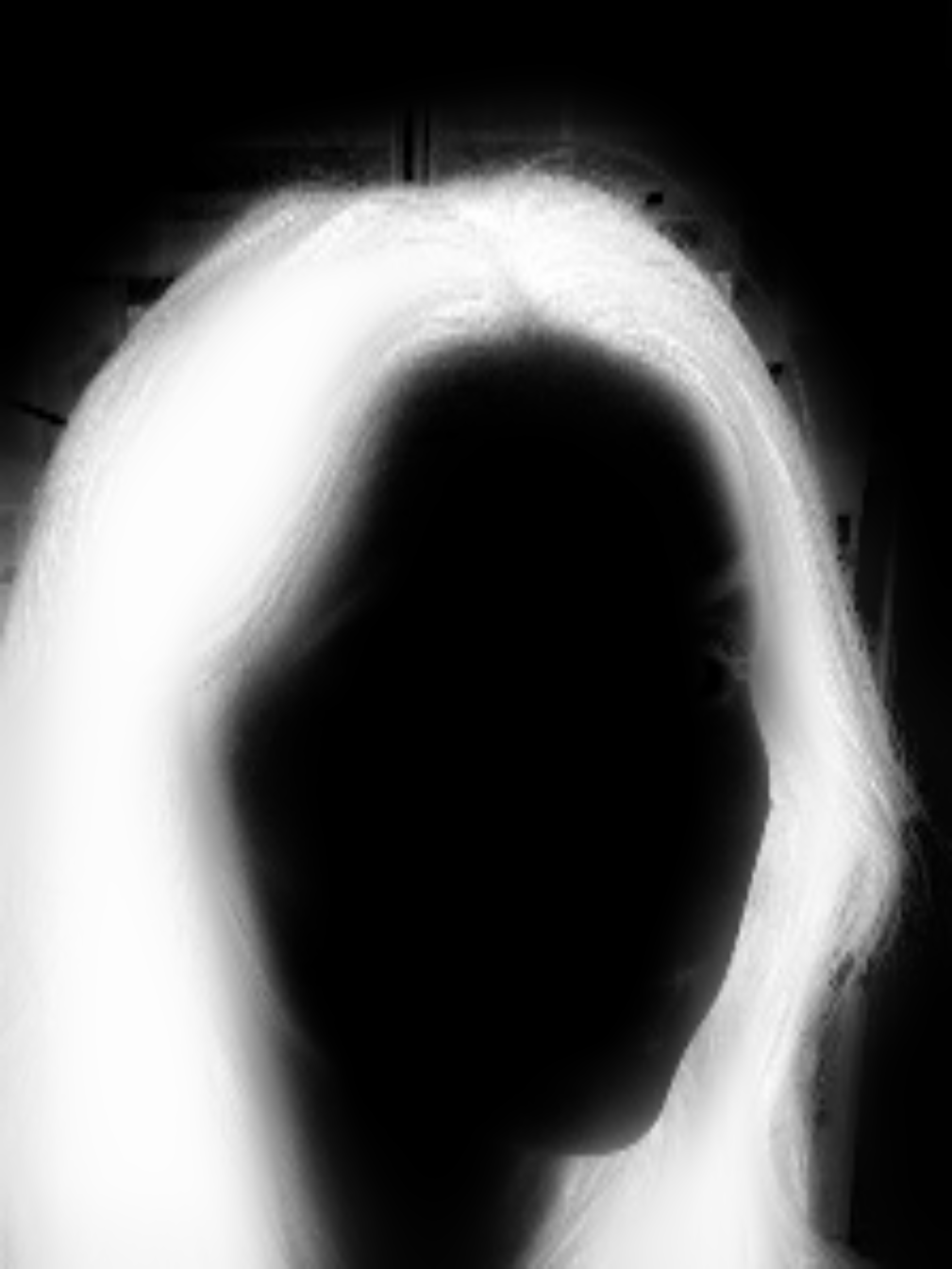}}
		{\includegraphics[width=0.24\textwidth]{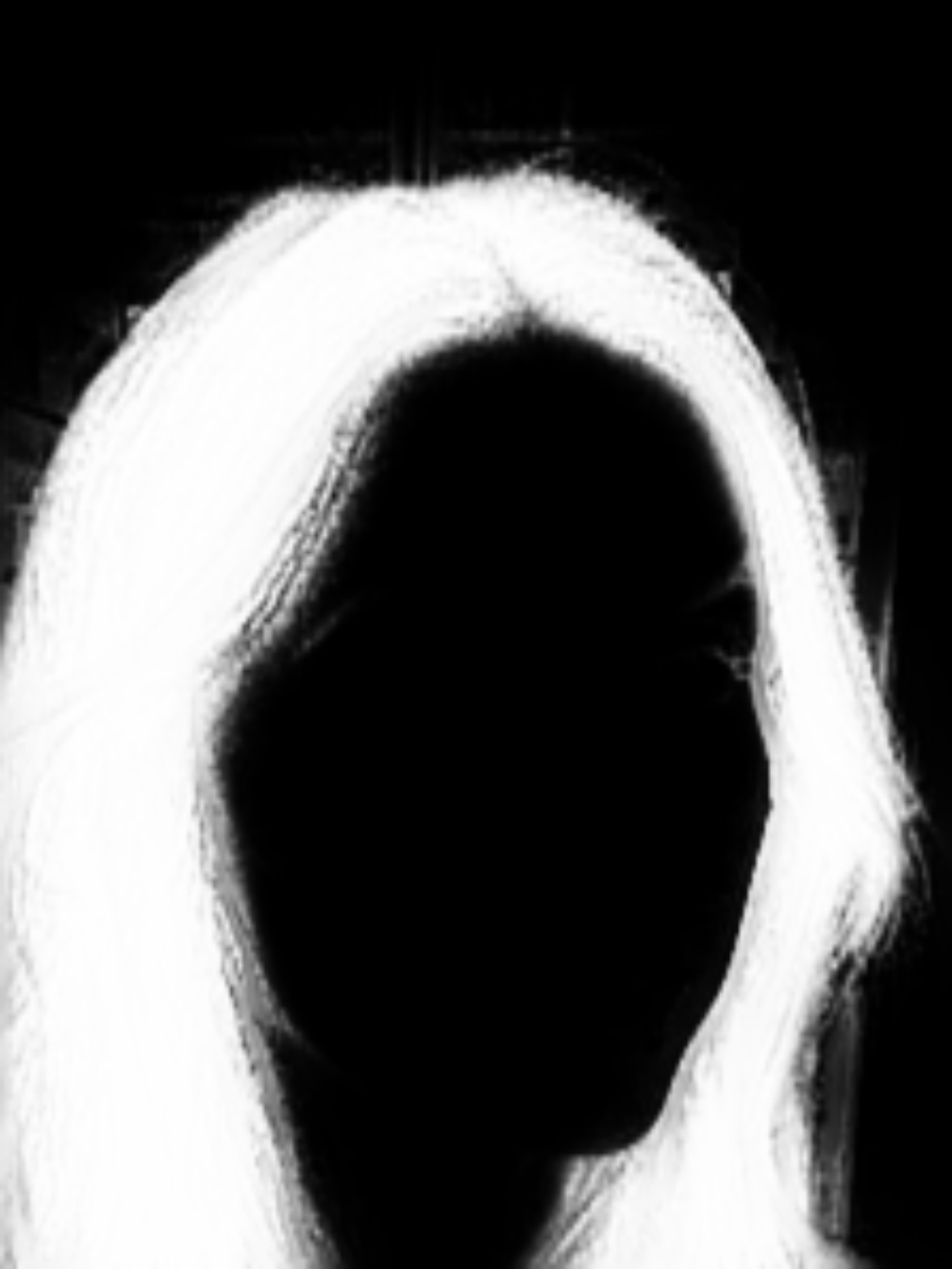}}}
	\vspace{3pt}
	\centerline{{\includegraphics[width=0.24\textwidth]{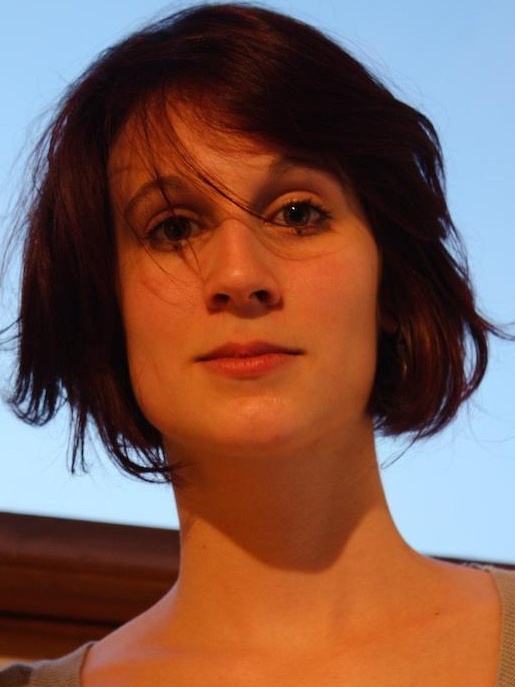}}
		{\includegraphics[width=0.24\textwidth]{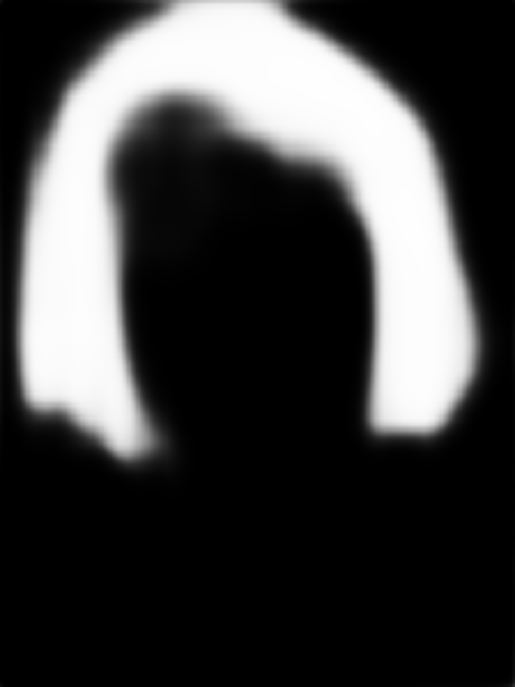}}
		{\includegraphics[width=0.24\textwidth]{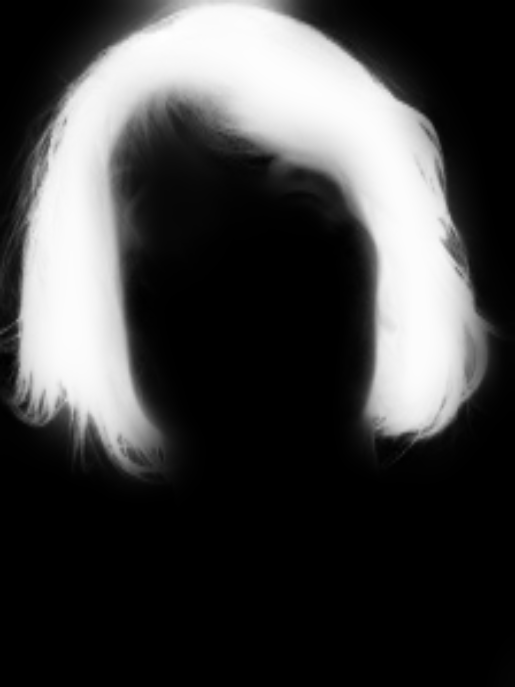}}
		{\includegraphics[width=0.24\textwidth]{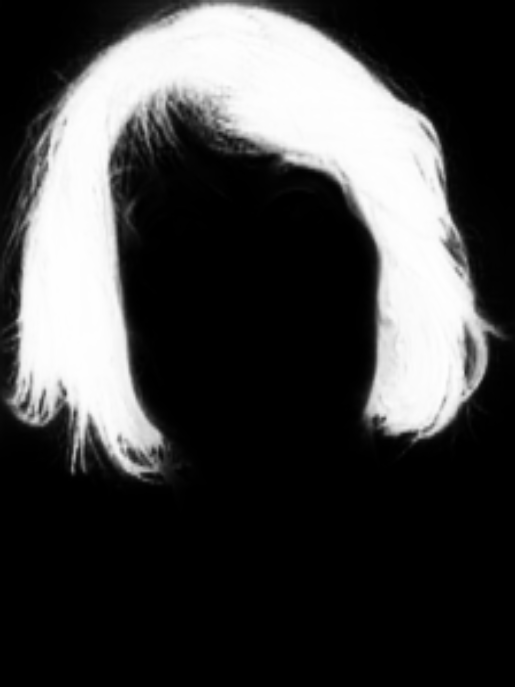}}}
	\vspace{3pt}
	\centerline{{\includegraphics[width=0.24\textwidth]{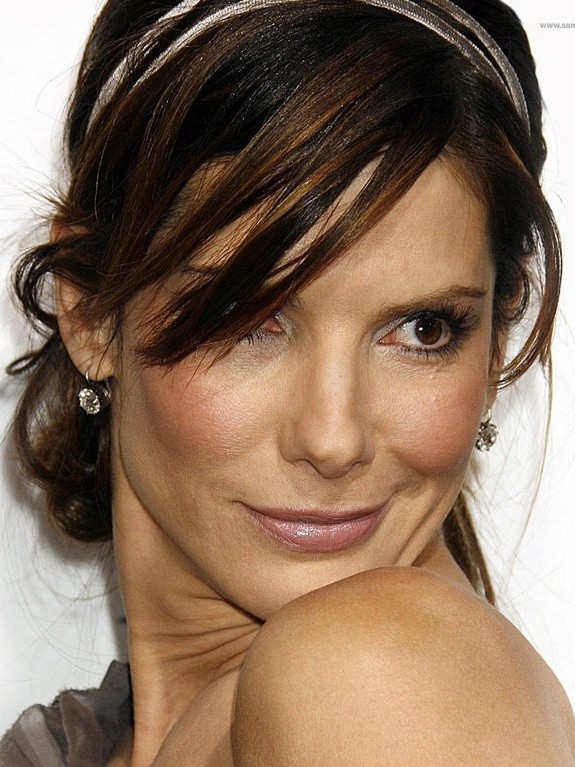}}
		{\includegraphics[width=0.24\textwidth]{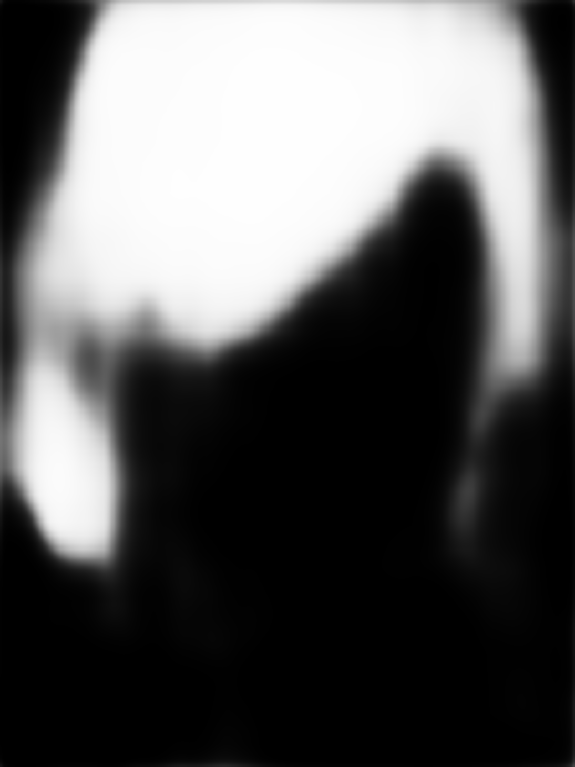}}
		{\includegraphics[width=0.24\textwidth]{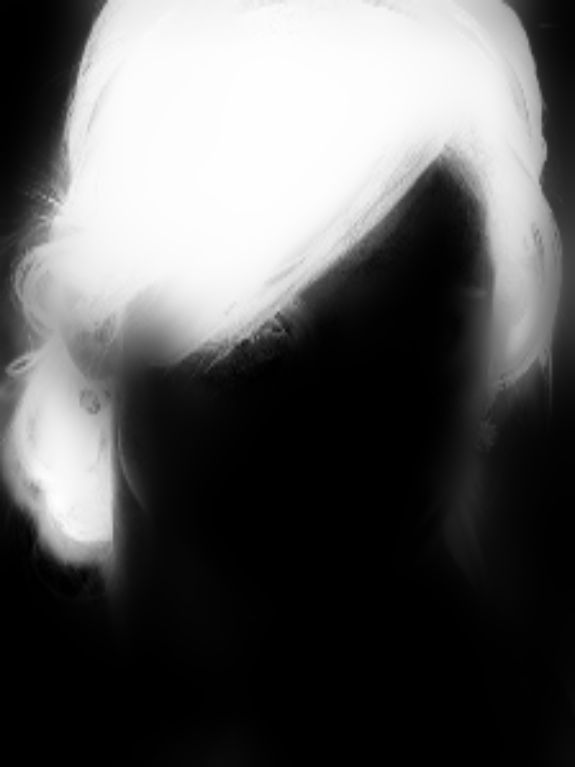}}
		{\includegraphics[width=0.24\textwidth]{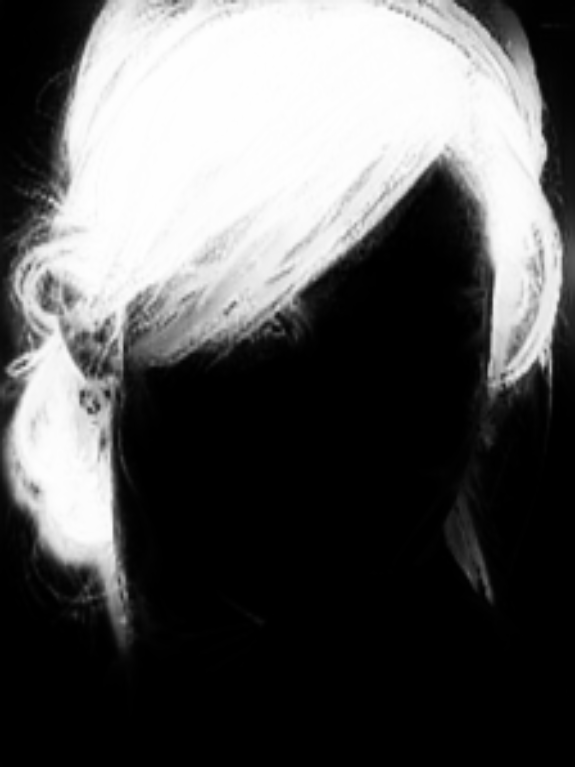}}}
	\vspace{3pt}
	\centerline{\subfigure[]{\includegraphics[width=0.24\textwidth]{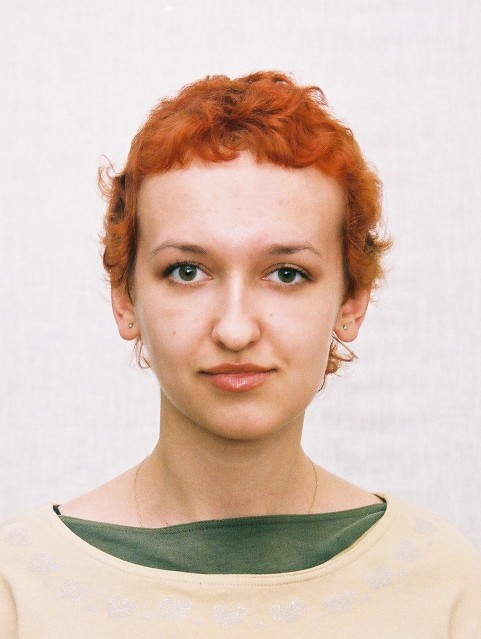}}
		\subfigure[]{\includegraphics[width=0.24\textwidth]{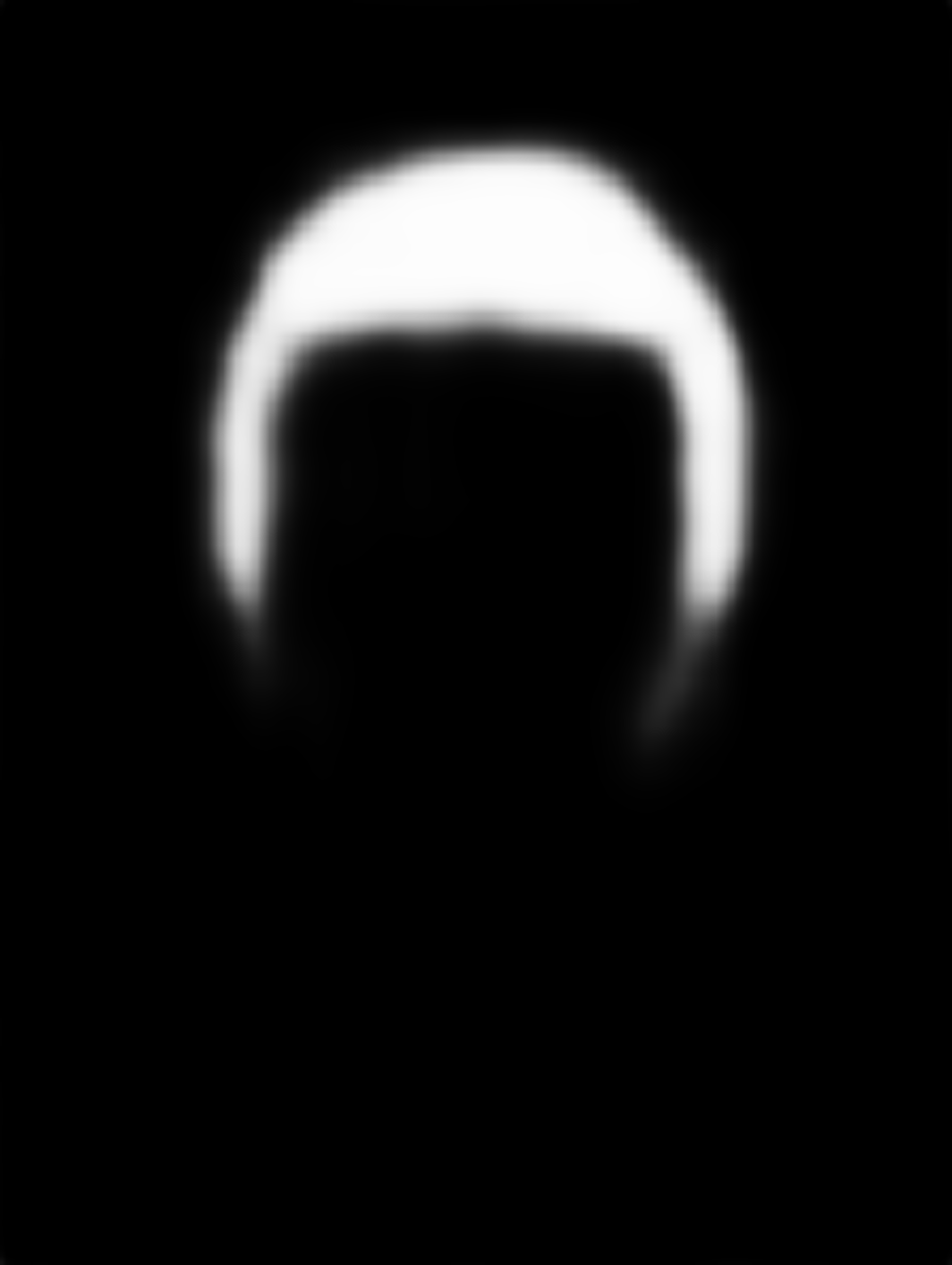}}
		\subfigure[]{\includegraphics[width=0.24\textwidth]{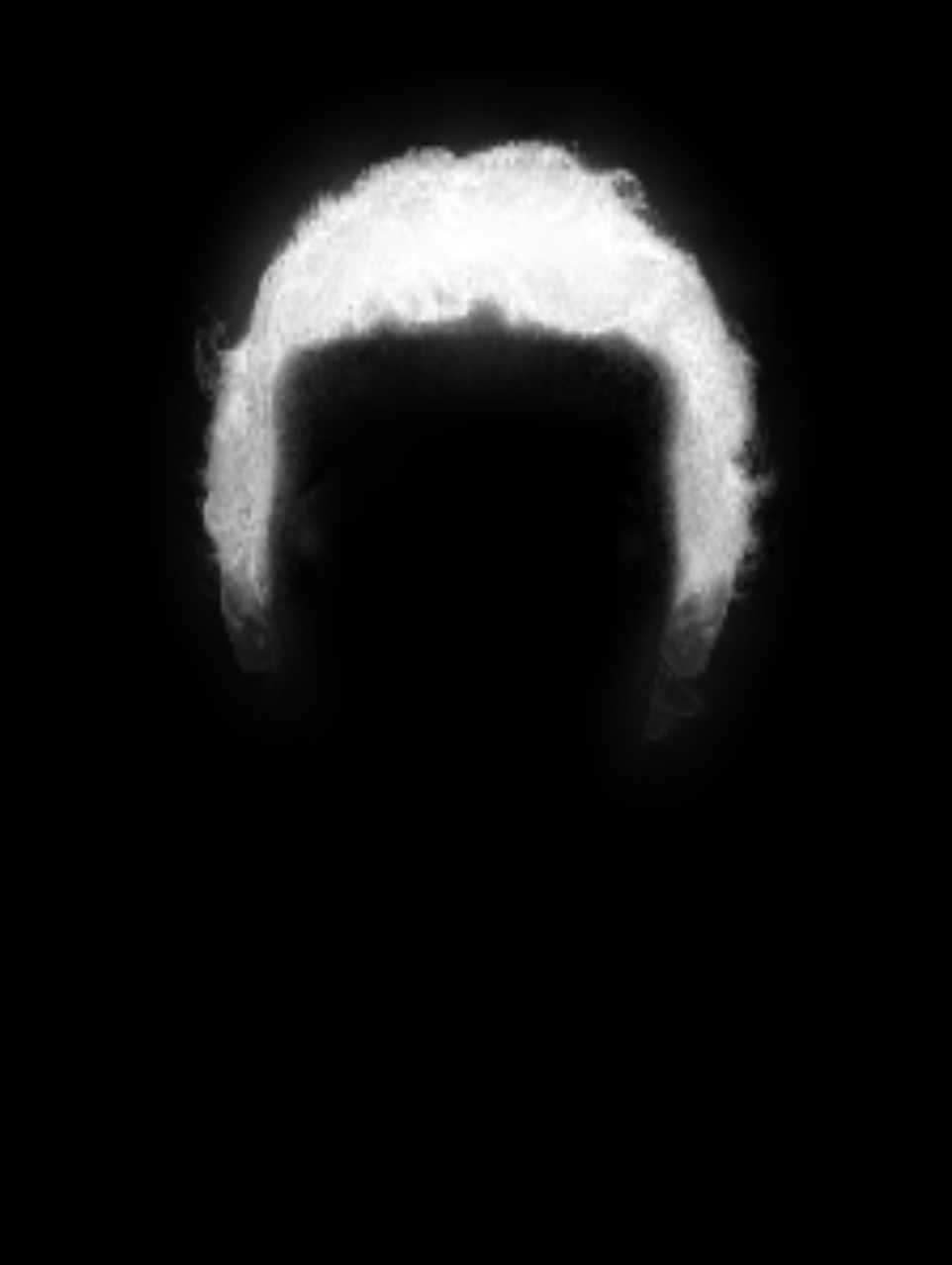}}
		\subfigure[]{\includegraphics[width=0.24\textwidth]{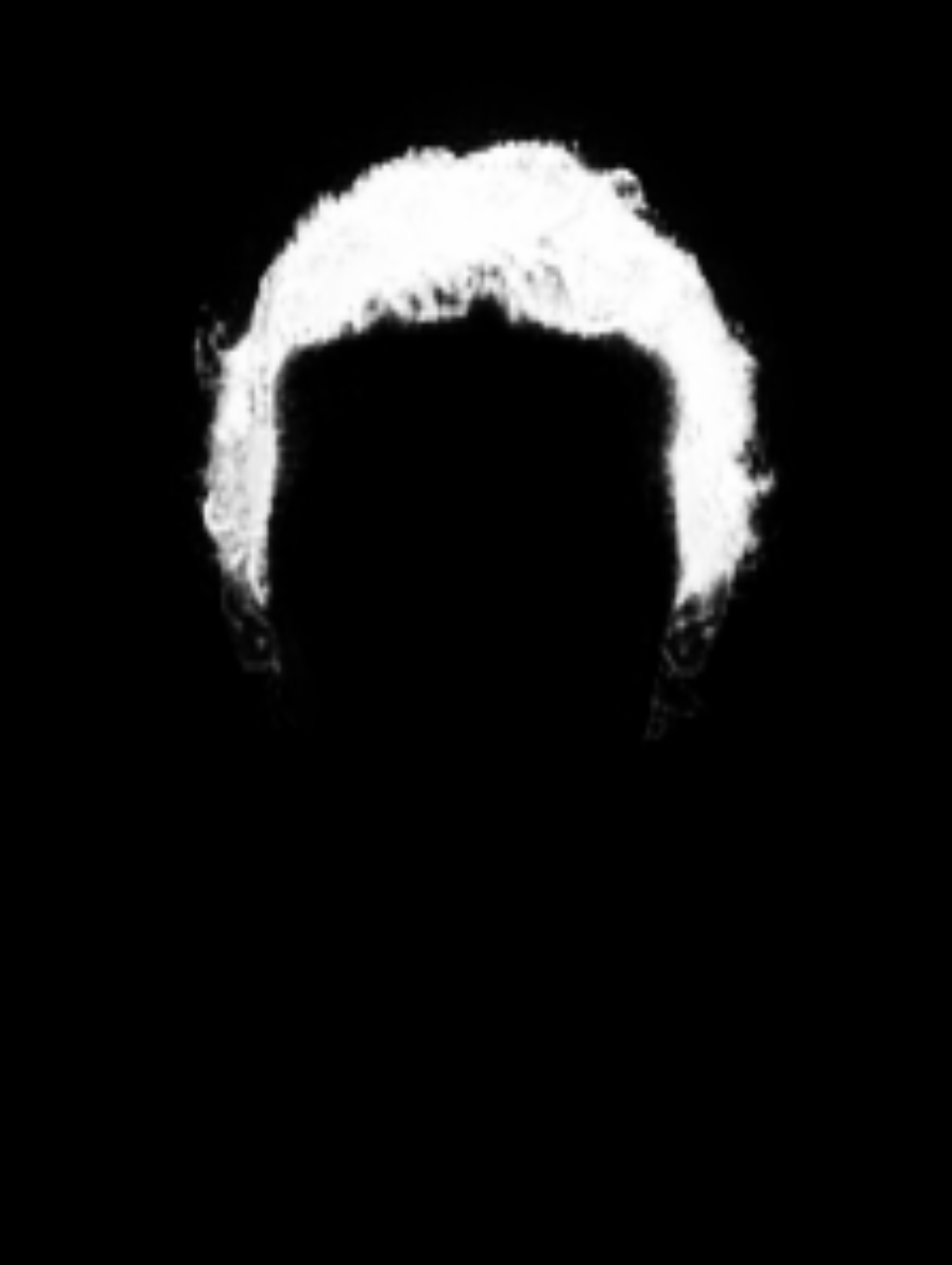}}}
	\end{minipage}
	\begin{minipage}{0.49\textwidth}
	\centerline{{\includegraphics[width=0.24\textwidth]{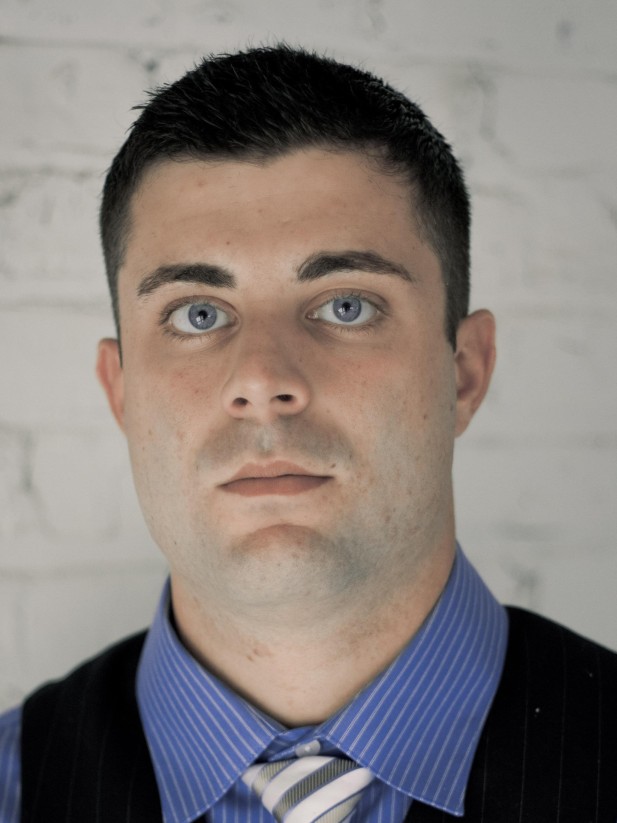}}
		{\includegraphics[width=0.24\textwidth]{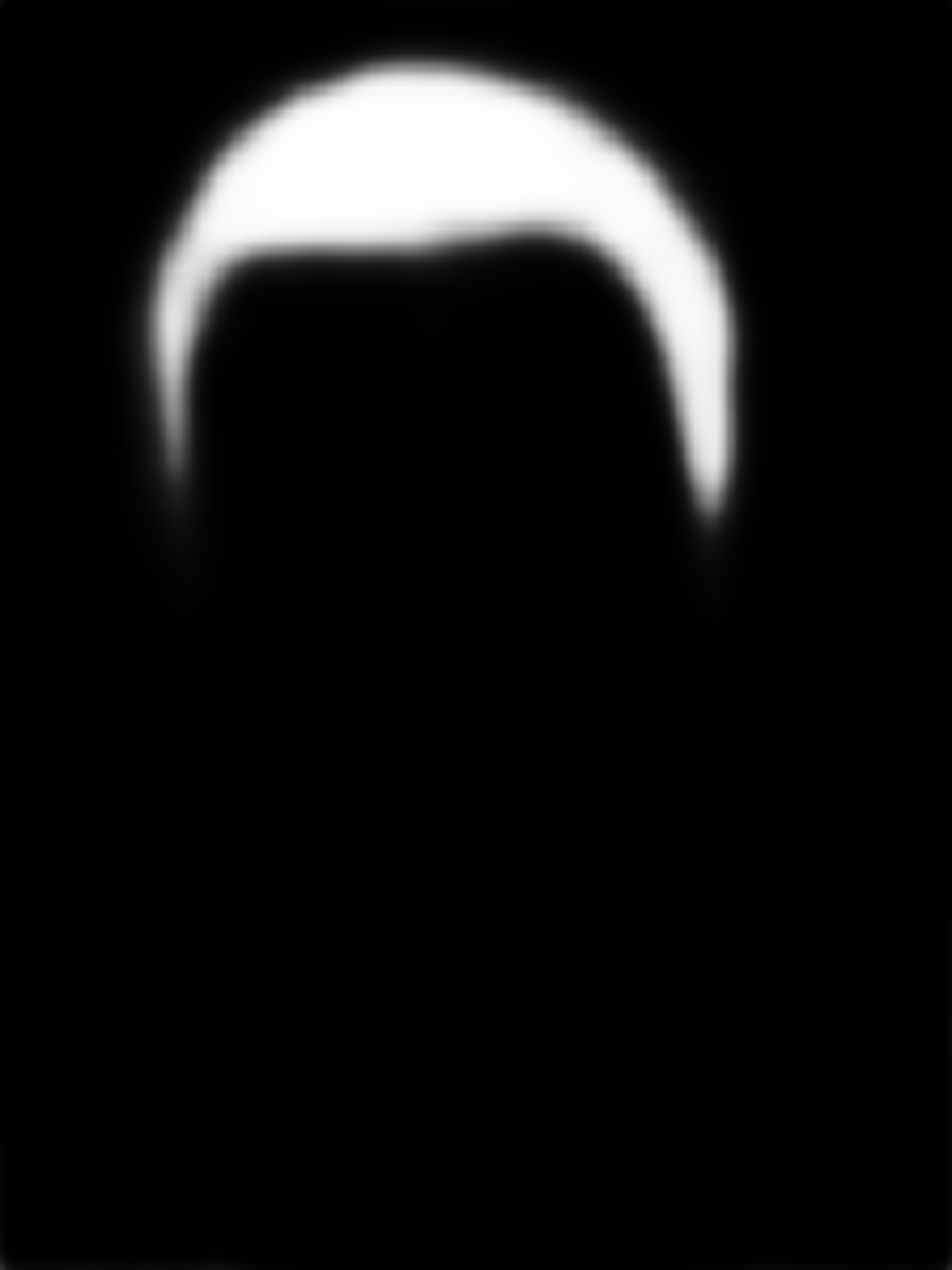}}
		{\includegraphics[width=0.24\textwidth]{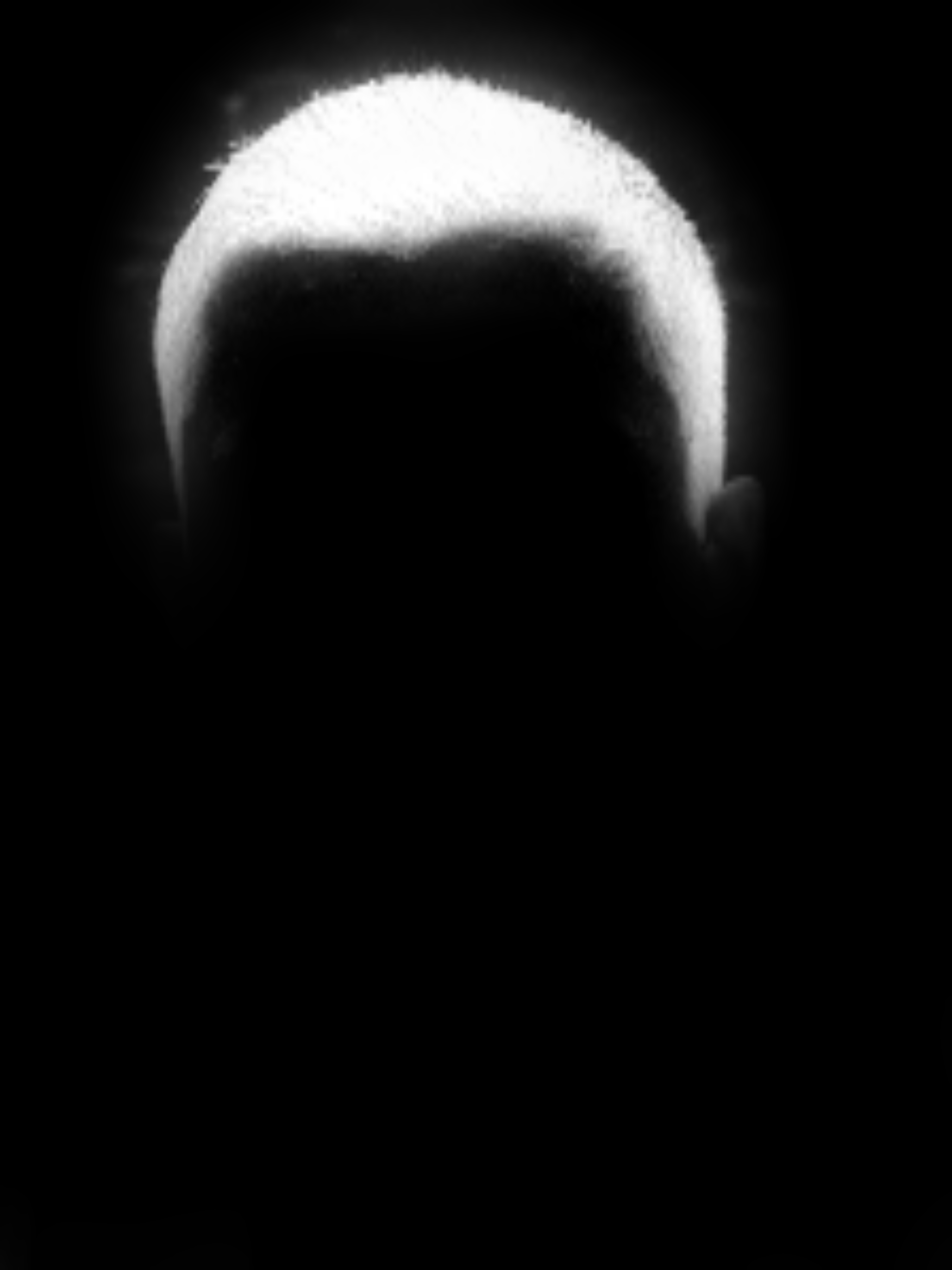}}
		{\includegraphics[width=0.24\textwidth]{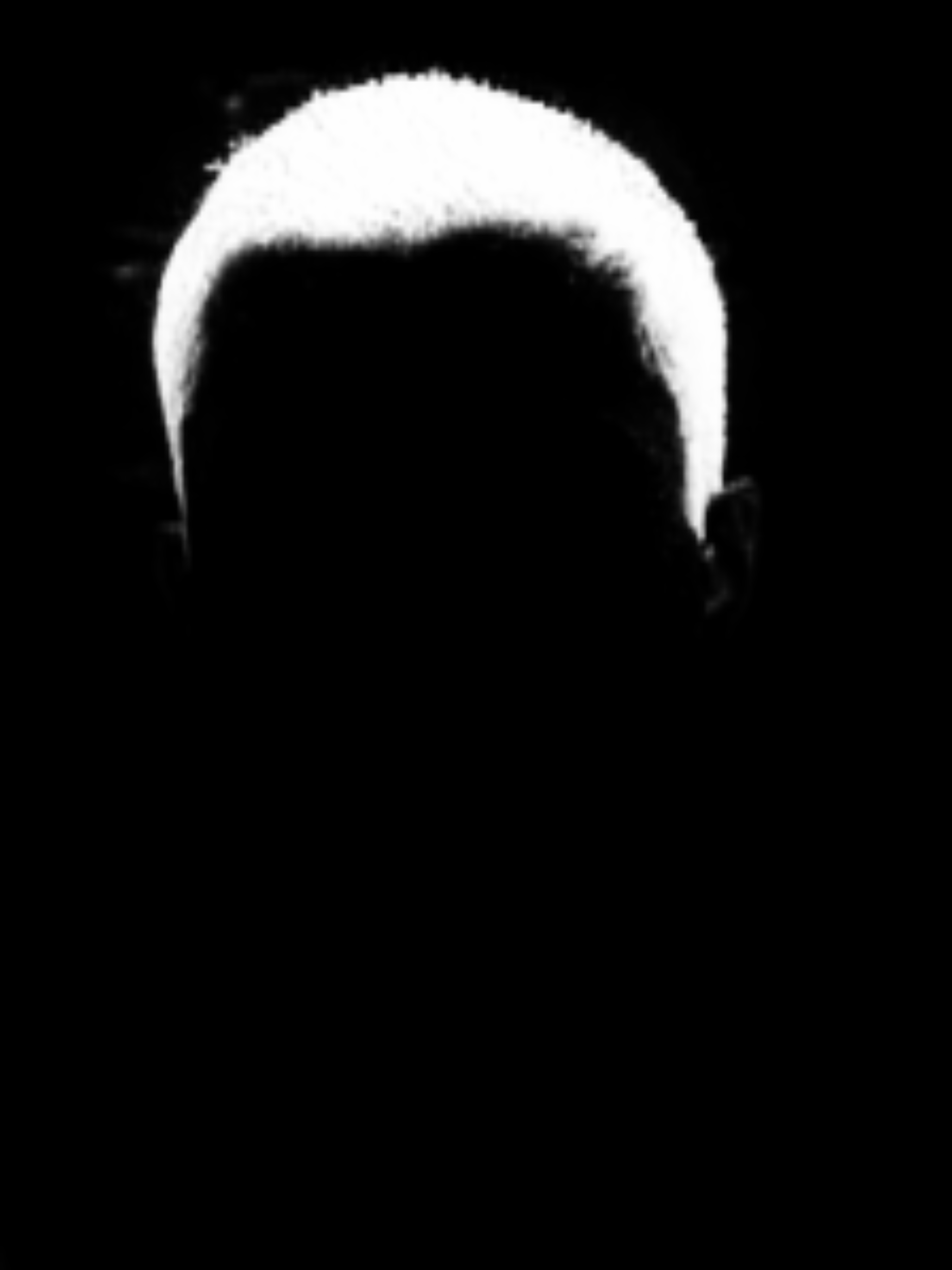}}}
	\vspace{3pt}	
	\centerline{{\includegraphics[width=0.24\textwidth]{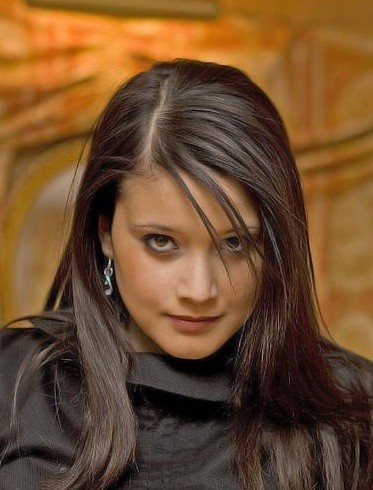}}
		{\includegraphics[width=0.24\textwidth]{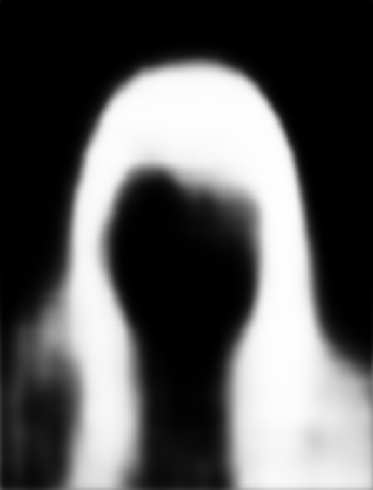}}
		{\includegraphics[width=0.24\textwidth]{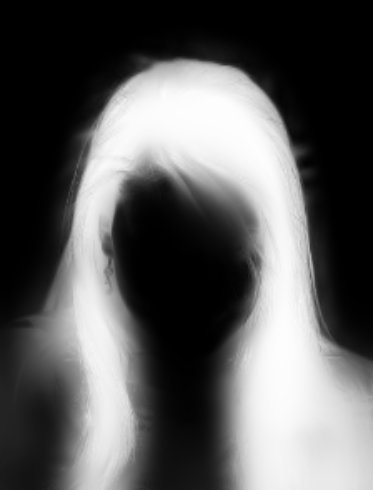}}
		{\includegraphics[width=0.24\textwidth]{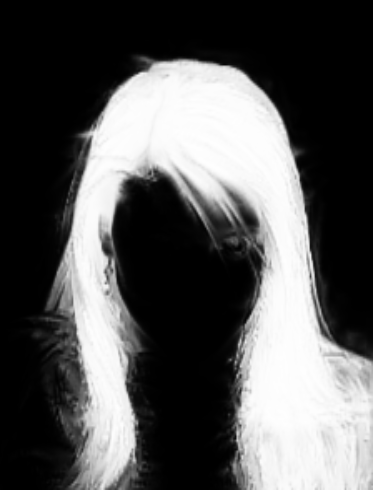}}}
	\vspace{3pt}	
	\centerline{{\includegraphics[width=0.24\textwidth]{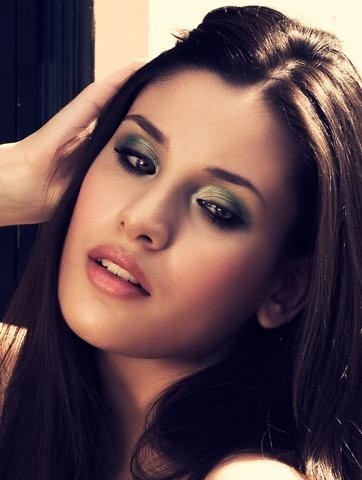}}
		{\includegraphics[width=0.24\textwidth]{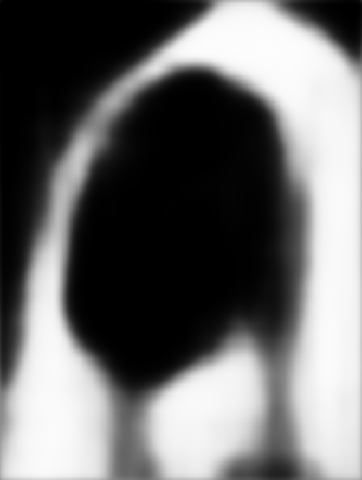}}
		{\includegraphics[width=0.24\textwidth]{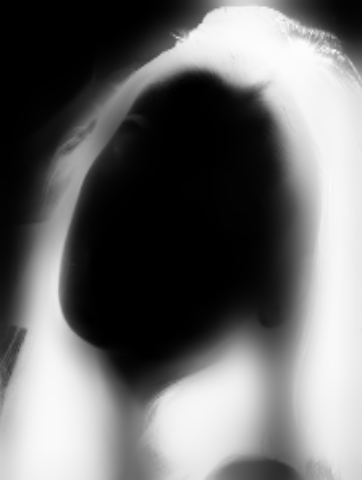}}
		{\includegraphics[width=0.24\textwidth]{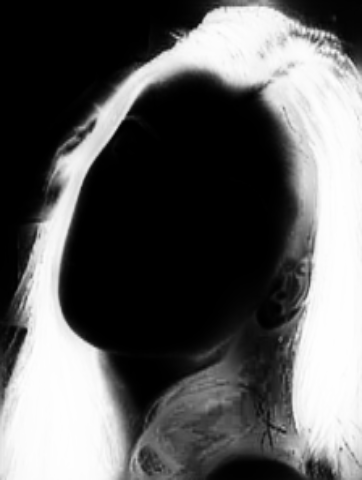}}}
	\vspace{3pt}	
	\centerline{\subfigure[]{\includegraphics[width=0.24\textwidth]{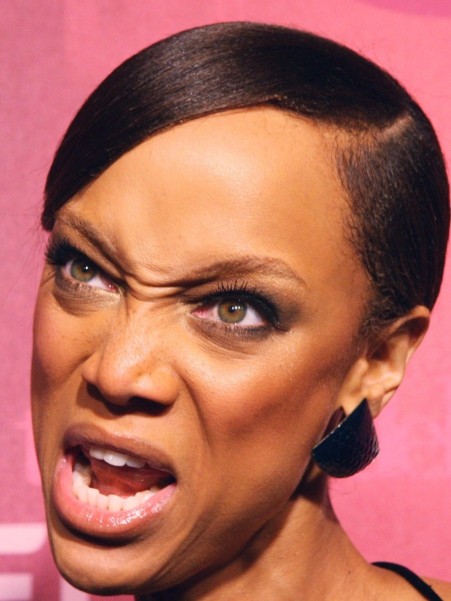}}
		\subfigure[]{\includegraphics[width=0.24\textwidth]{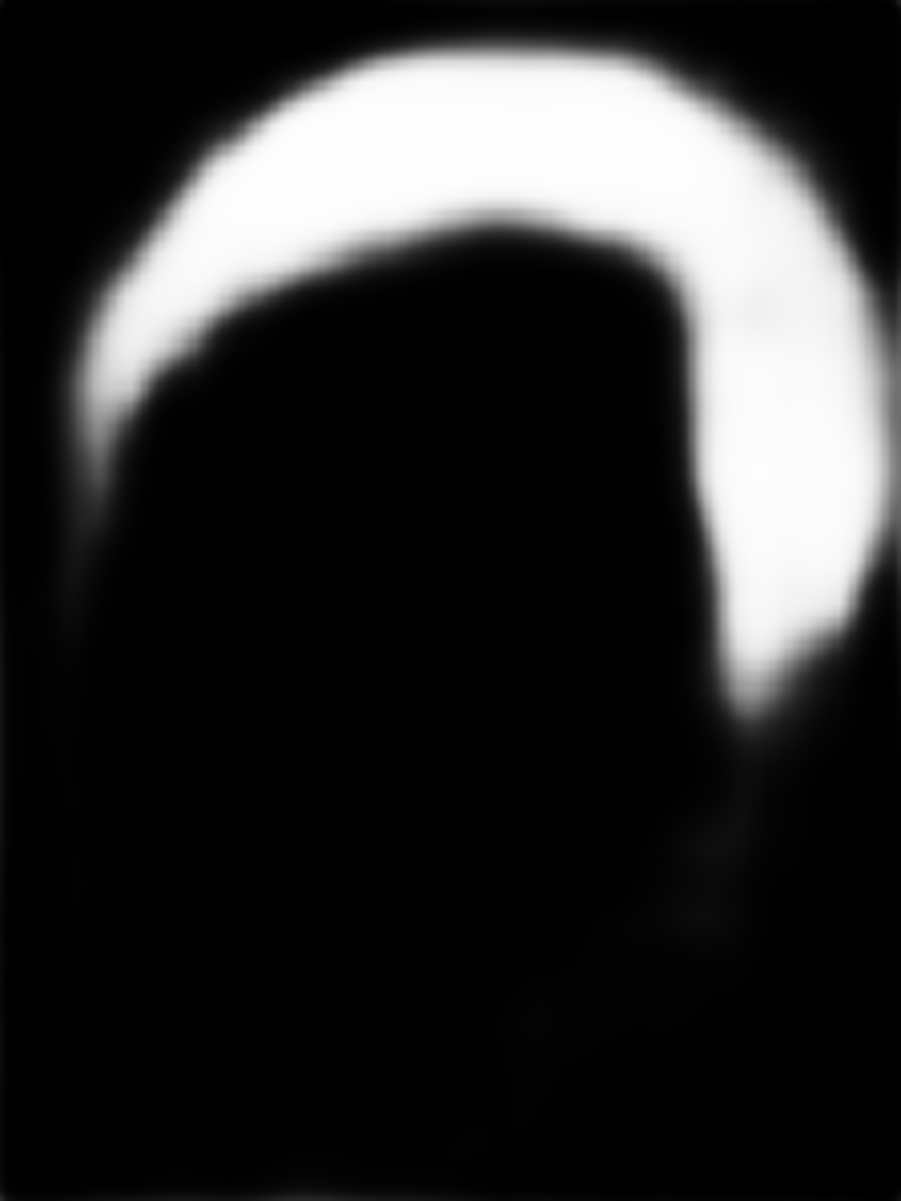}}
		\subfigure[]{\includegraphics[width=0.24\textwidth]{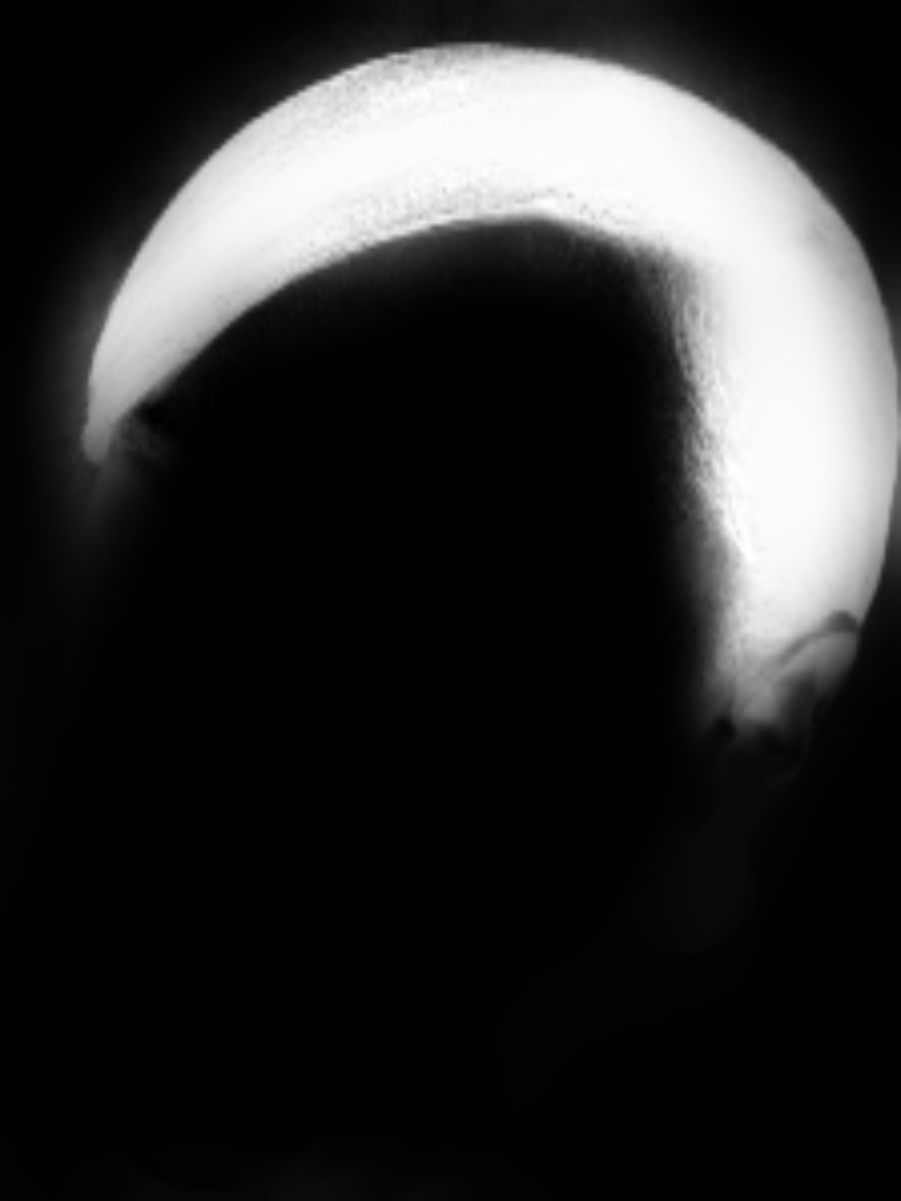}}
		\subfigure[]{\includegraphics[width=0.24\textwidth]{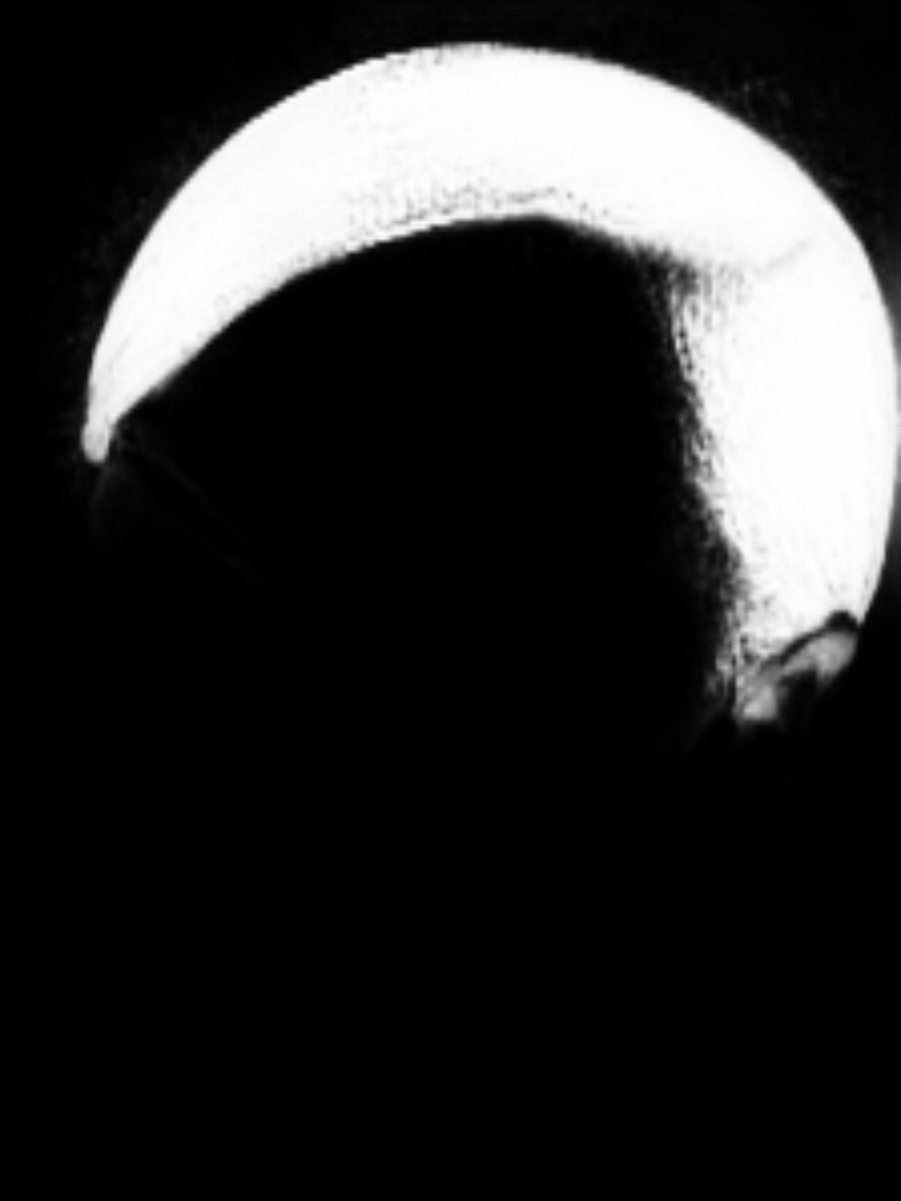}}}
	\end{minipage}}
	\caption{Qualitative evaluation. (a,e) Input image. (b,f) HairSegNet. (c,g) HairSegNet + Guided Filter. (d,h) HairMatteNet.}
	\label{fig:qualitative}
\end{figure*}

\begin{table}[!t]
	\footnotesize
	\centering
	\setlength\tabcolsep{3pt}
	\begin{tabular}{|c|c|c|c|c|c|}
		\hline
		Model &  F1 & Perf. & IoU & Acc. & Grad-cons.\\
		\hline
		\multicolumn{6}{|c|}{Crowd-sourced dataset} \\
		\hline
		HairSegNet &  $0.9212$ & $0.7833$ & $0.8564$ & $0.9624$ & $0.2696$ \\
		\hline
		HairSegNet + GF &  $0.9251$ & $0.7964$ & $0.8637$ & $0.9646$ & $0.1803$ \\
		\hline
		HairMatteNet &  $0.9219$ & $0.7945$ & $0.8589$ & $0.9619$ & $0.0533$ \\
		\hline
		\multicolumn{6}{|c|}{LFW Parts dataset \cite{GLOC_CVPR13}} \\
		\hline
		HairMatteNet &  $NA$ & $NA$ & $NA$ & $0.965$ & $NA$ \\
		\hline
		\cite{qin2017automatic} FCN+CRF  &  $NA$ & $NA$ & $NA$ & $0.9676$ & $NA$ \\
		\hline
		HairMatteNet SP &  $NA$ & $NA$ & $NA$ & $0.9769$ & $NA$ \\
		\hline
		\cite{qin2017automatic} FCN+CRF SP &  $NA$ & $NA$ & $NA$ & $0.9732$ & $NA$ \\
		\hline
		\multicolumn{6}{|c|}{Guo and Aarabi dataset \cite{guo2016hair}} \\
		\hline
		HairMatteNet &  $0.9376$ & $0.8253$ & $0.8848$ & $0.964$ & $0.0532$ \\
		\hline
		HNN \cite{guo2016hair} &  $0.7673$ & $0.4674$ & $0.6454$ & $0.8793$ & $0.2732$ \\
		\hline
	\end{tabular}
	\caption{Quantitative evaluation}
	\label{tab:crowd_eval}
\end{table}

\subsection{Qualitative evaluation}
We evaluate our method on publicly available selfie images for qualitative analysis. Results can be seen in Fig.~\ref{fig:qualitative}. HairSegNet (Fig.~\ref{fig:qualitative}b) yields good but coarse masks. HairSegNet with Guided Filter (Fig.~\ref{fig:qualitative}c) produces better masks but with an undesirable blur around hair boundaries. The most accurate and sharpest results are achieved by HairMatteNet (Fig.~\ref{fig:qualitative}d). A failure mode of both Guided Filter post-processing and HairMatteNet is their under-segmentation of hair-like objects in the vicinity of hair, such as eyebrows in case of dark hair or bright background for light hair. In addition, highlights inside the hair can cause the hair mask from HairMatteNet to be non-homogeneous, which is especially apparent in the last three examples in column (h).

\subsection{Network architecture experiments}
\noindent
{\bf Decoder layer depth } Using our validation data, we have experimented with number of decoder layer channels, but observed that it does not have a large effect on accuracy. Table \ref{tab:decoder_depth} illustrates our experiments with the number of channels in the decoder, with $64$ channels yielding the best results according to most measures. These experiments were done using the skip connections architecture in Fig. \ref{fig:mobilenet_segmentation_skip} without using the gradient consistency loss.

\begin{table}[!t]
	\footnotesize
	\centering
	\setlength\tabcolsep{3pt}
	\begin{tabular}{|c|c|c|c|c|}
		\hline
		Depth &  F1 & Perf. & IoU & Acc.\\
		\hline
		16 &  $0.9202$ & $0.7878$ & $0.8559$ & $0.9585$ \\
		\hline
		32 &  $0.9217$ & $0.8001$ & $0.8584$ & $0.9581$ \\
		\hline
		64 &  $0.9229$ & $0.7939$ & $0.8608$ & $0.9604$ \\
		\hline
		128 & $0.9225$ & $0.7937$ & $0.8605$ & $0.9596$ \\
		\hline
	\end{tabular}
	\caption{Decoder layer depth experiments on validation data}
	\label{tab:decoder_depth}
\end{table}

\vspace{4pt}

\begin{figure}[!t]
	\centerline{{\includegraphics[width=0.24\textwidth]{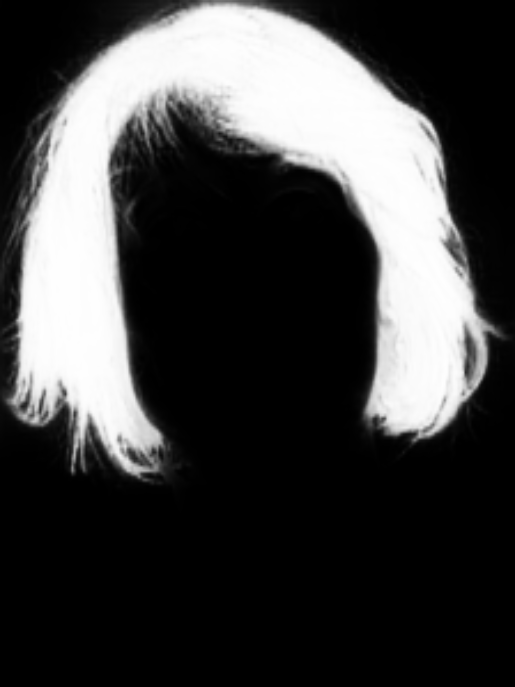}}
		{\includegraphics[width=0.24\textwidth]{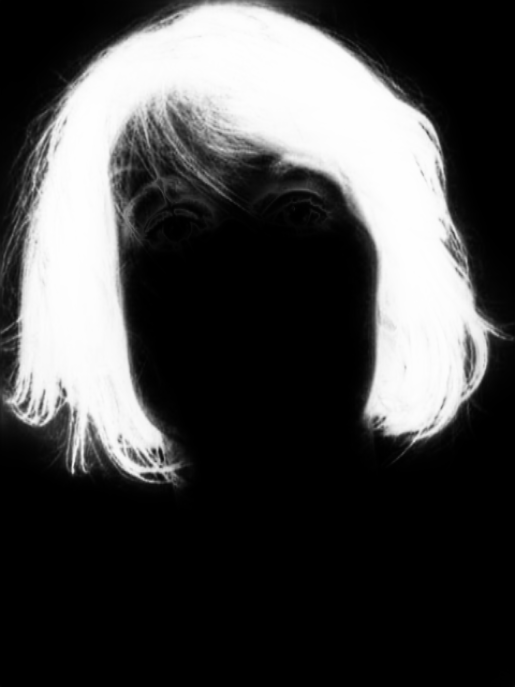}}
		}
	\vspace{3pt}
	\centerline{\subfigure[]{\includegraphics[width=0.24\textwidth]{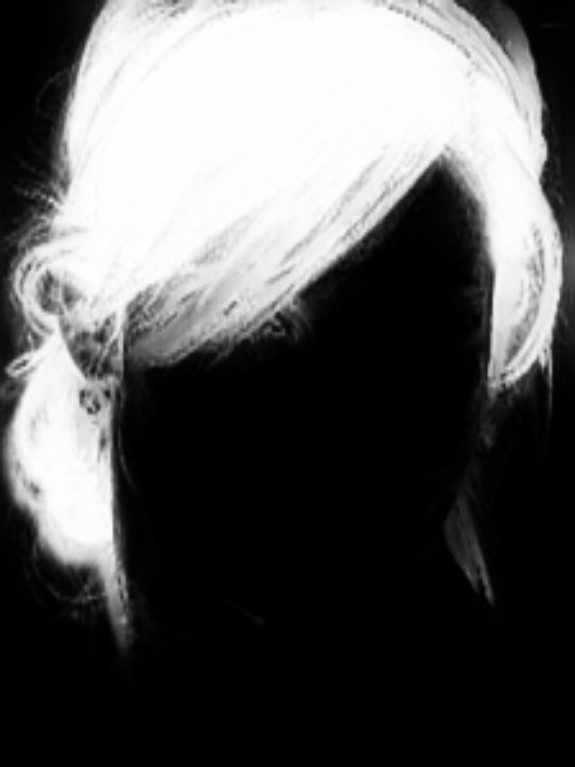}}
		\subfigure[]{\includegraphics[width=0.24\textwidth]{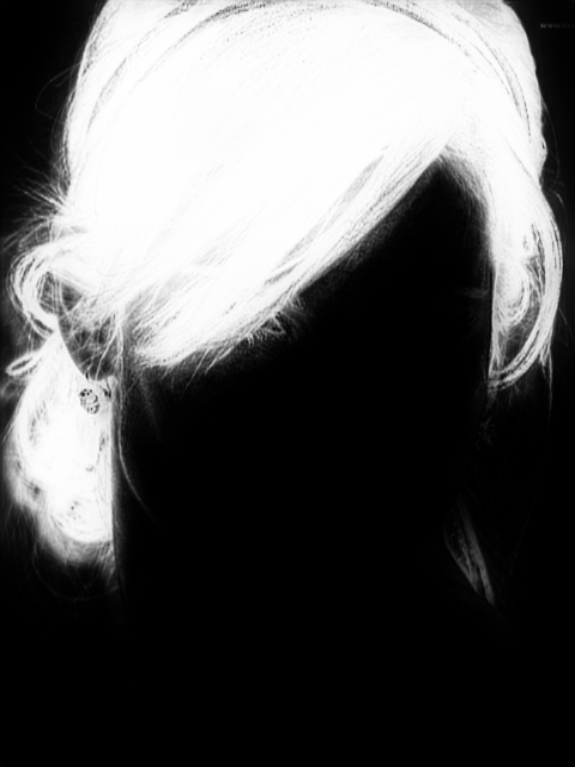}}
	}
	\caption{Network resolution comparison (a) $224 \times 224$ (b) $480 \times 480$}
	\label{fig:res_comp}
\end{figure}

\noindent
{\bf Input image size } Howard et al. \cite{howard2017mobilenets} observed that MobileNets perform better given higher image resolution. Given our goal of accurate hair matting, we experimented with increasing the resolution beyond $224 \times 224$, which is the highest resolution MobileNet were trained on ImageNet. For the 2nd and 3rd image in Fig. \ref{fig:qualitative}, Fig. \ref{fig:res_comp} shows qualitative comparison of masks inferred using HairMatteNet from $224 \times 224$ images vs. $480 \times 480$ images. The $480 \times 480$ results look more accurate around the hair edges, with longer hair strands being captured (e.g., the long hair strand falling on the nose in the first image). However, the issues mentioned in the previous section are emphasized as well, with more of the hair mask bleeding into non-hair regions and the inside of the mask becoming non-homogeneous due to hair highlights. In addition, processing a larger image is significantly more expensive.

\section{Summary}
\label{sec:summary}
This paper presented a hair matting method with real-time performance on mobile devices. We have shown how, given noisy and coarse data, a modified MobileNet architecture is trained to yield accurate matting results. While we apply the proposed architecture for hair matting, it is general and can be applied to other segmentation tasks. In future work we will explore fully automatic methods for training from noisy data without the need for manual filtering. In addition, we will explore further improvements to matting quality, such as capturing longer hair strands and segmenting light hair; all this while keeping the hair mask homogeneous, preventing it from bleeding into non-hair regions, and maintaining our real-time performance on mobile devices.




\bibliographystyle{IEEEtran}
\bibliography{HairSegmentation_2017}

\end{document}